\documentclass[twoside]{article}
\usepackage{graphicx}
\usepackage{array}
\usepackage{lipsum} 
\usepackage[sc]{mathpazo} 
\usepackage[T1]{fontenc} 
\usepackage{microtype} 
\usepackage[hmarginratio=1:1,top=32mm,columnsep=20pt]{geometry} 
\usepackage{multicol} 
\usepackage[hang, small,labelfont=bf,up,textfont=it,up]{caption} 
\usepackage{booktabs} 
\usepackage{float} 
\usepackage{hyperref} 
\usepackage{lettrine} 
\usepackage{paralist} 
\usepackage{abstract} 
\usepackage{multicol}
\usepackage{lipsum}

\usepackage{titlesec} 
\titleformat{\section}[block]{\large\scshape\centering}{\thesection.}{1em}{} 
\titleformat{\subsection}[block]{\large}{\thesubsection.}{1em}{} 
\usepackage{fancyhdr} 
\title{\vspace{-15mm}\fontsize{24pt}{10pt}\selectfont\textbf{ Tracking Individual Targets in High Density Crowd Scenes\\
Analysis of a Video Recording in Hajj2009}} 

\author{
\large
\textsc{Mohamed H. Dridi}\\
\normalsize Institute of Theoretical Physics 1\\
University of Stuttgart \\
\normalsize \href{mailto:mohamed.dridi@itp1.uni-stuttgart.de}{mohamed.dridi@itp1.uni-stuttgart.de} 
\vspace{5mm}}

\date{3/07/2014} 

\begin{document}
\maketitle 
\thispagestyle{fancy}
%----------------------------------------------------------------------------------------
%	ABSTRACT
%----------------------------------------------------------------------------------------

\begin{abstract}
\noindent In this paper we present a number of methods (manual, semi-automatic and automatic) for tracking individual targets in high density crowd scenes where thousand of people are gathered. The necessary data about the motion of individuals and a lot of other physical  information can be extracted from consecutive image sequences in different ways, including optical flow and block motion estimation. One of the famous methods for tracking moving objects is the block matching 
method. This way to estimate subject motion requires the specification of a comparison window which determines the scale of the estimate.
In this work we present a real-time method for pedestrian recognition and tracking in sequences of high resolution images obtained by a stationary (high definition) camera located in different places on the Haram mosque in Mecca. The objective is to estimate pedestrian velocities as a function of the local density.The resulting data of tracking moving pedestrians based on video sequences are presented in the following section. Through the evaluated system  the spatio-temporal coordinates of each pedestrian during the Tawaf ritual are established. The pilgrim velocities as function of the local densities in the Mataf area (Haram Mosque Mecca) are illustrated and very precisely documented. 

Tracking in such places where pedestrian density reaches 7 to 8 Persons/m$^2$ is extremely challenging due to the small number
of pixels on the target,  appearance ambiguity resulting from the dense packing, and severe inter-object occlusions. The tracking method which is outlined in this paper overcomes these challenges by using a virtual camera which is matched in position, rotation and focal length to the original camera in such a way that the features of the 3D-model match the feature position of the filmed mosque. In this model an individual feature has to be identified by eye, where contrast is a criterion. We do know that the pilgrims walk on a plane, and after matching the camera we also have the height of the plane in 3D-space from our 3D-model. A point object is placed at the position of a selected pedestrian. During the animation we set multiple animation-keys (approximately every 25 to 50 frames which equals 1 to 2 seconds) for the position, such that the position of the point and the pedestrian overlay nearly at every time.
By combining all these variables with the available appearance information, we are able to track individual targets in high density crowds. 

\end{abstract}

%----------------------------------------------------------------------------------------
%	ARTICLE CONTENTS
%----------------------------------------------------------------------------------------

\begin{multicols}{2} % Two-column layout throughout the main article text

\section{Introduction}

\lettrine[nindent=0em,lines=3]{C} rowd simulation has found its way into computer science, computer visualizations and the computer simulation of oriented building construction and crowd management \cite{Predtetschenski-Milinski1969}. 
With continuously growing population around the world and with enormous evolution in the different modes of transportation in the last decade a lot of paper have appeared with increasing interest in modelling crowd and evacuation dynamics. Thus the simulation of pedestrian flows has become an important research area. Pedestrian models are based on macroscopic or microscopic behaviour. 

The evolution and design of any pedestrian simulation model requires a lot of information and data.

A number of variables and attributes arises from empirical data collection and need to be considered to develop and calibrate a (microscopic) pedestrian simulation model. 

For this reason we used different tools and developed different methods to collect the microscopic data and to
analyse microscopic pedestrian flow. It is very important to mention that  the pedestrian data collection especially in a dangerous situation is still very much in its infancy. An aim of this study is to establish more clearness and understanding about the microscopic pedestrian flow characteristics. Manual, semi manual and automatic image processing data collection systems were developed. Many published studies show that the microscopic speed obey a normal distribution with a mean of 1.38 m/second and a standard deviation of 0.37 m/second. The acceleration distribution also resemblances a normal distribution with an average of 0.68 m/ square second \cite{Knoblauch1996, Fruin1971a, o1996transport}.

For the evolution and development of pedestrian microscopic simulation models, a lot of data was collected with the help of video recording and tracking of moving entities in the pedestrian flow using the coordinates of the head path was established through image processing. A large trajectory dataset has been restored. For the observation of pedestrian flows in public places a Sony camera was used. This observation was in different places where the pilgrims perform their rituals. Many variables can be gathered to describe the behaviour of pedestrians from different points of view.
This paper describes how to obtain variables from video taking and simple image processing that can represent the movement of pedestrians (pilgrims) and its variables. Moreover in this work we try to understand several parameters influencing the pedestrian behaviour in riots or panic situations. 

For obtaining empirical data different methods were used, automatic and manual methods. We have analysed video recordings of the crowd movement in the Tawaf in Mosque/Mecca during the Hajj on the 27th of November, 2009. We have evaluated unique video recordings of a 105$\times$ 154 m large Mataf area taken from the roof of the Mosque, where upto 3 million Muslims perform the Tawaf and Sa'y rituals within 24 hours.

Both Microscopic Video Data Collection and Microscopic Pedestrian Simulation Model generate a database called PedFlow database. The properties and characteristics that are capable of explaining microscopic pedestrian flow are illustrated. A comparison between average instantaneous speed distributions describing the real world obtained from different methods, and how they can be used in the calibration and validation of the simulation tools, are explained.

\section{Related work}
Typically, manual counting was performed by tally sheet or mechanical or electronic count board to collect density and speed data for pedestrian. Pedestrian behaviour studies are collected by manual observation or video recording in different public places like corridors side walks and cross walks. The effectiveness of the data (pedestrian speed) collected on any observed area is strongly related to the number of pedestrians in the flow. The relationship between speed, flow, and pedestrian density for a crowd population or human group has been published in many fundamental diagrams developed by Fruin \cite{Fruin1987} and others \cite{Seyfried2009}. Though for many reasons the method has been used to detect and count vehicles in automatic way cannot be used to detect pedestrians, since this system has been evaluated through pneumatic tube or inductance loops. As we can deduce from later work on this technology the possibility of applying this method to reproduce trajectory and motion prediction is still in a discussion phase. 

Other approaches use a neural network framework recursively to predict pedestrian motion and trajectory \cite{Bulpitt1998}. However the pedestrian trajectories in this system are calculated with incorrect simplifications. In particular, only the nearest neighbour trajectories are considered. The main  shortcoming of such an estimation is that there is no uncertainty in this prediction, moreover a comparison of different path prediction shows this is still far from the reality in order to predict that all objects will follow the same set of paths exactly.

A method which allowed people counting based on video texture synthesis and to reproduce motion in a novel way was introduced by Heisele and Woehler \cite{heisele1998}. The method works under the  assumption that people can be segmented from the moving background by means of appearance or motion properties. The scene image is clustered based on the color and position (R, G, B, X, Y) of pixel. The appearance of each pixel in a video frame is modelled as a mixture of Gaussian distributions. A algorithm is used that matches a spherical crust template to the foreground regions of the depth map. Matching is done by a time delay neural network for object recognition and motion analysis.

A significant task in video intelligence systems is the extraction of information about a moving objects e.g. detecting a moving crowd with  PedCount (a pedestrian counter system using CCTV) was developed by Tsuchikawa \cite{Tsuchikawa1995}. It extracts the object using the one line path in the image by background subtraction to make a space-time
(X-T) binary image. The direction of each travelling pedestrian is realized by the attitude of pedestrian region in the X-T image. They reported the need of background image reconstruction due to image illumination change. An algorithm to distinguish moving object from illumination change is explained based on the variance of the pixel value and frame difference. 

\section{Analyse of the Video Taking in Hajj2009}
\subsection{Introduction}
The electronic and digital revolution in video techniques during recent years has made it possible to gather detailed data concerning pedestrian behaviour, both in experiments and in real life situations \cite{Hoogendoorn2005, Johansson2008, Boltes2010}. 
The big challenge is to develop a new efficient method of defining and measuring basic quantities like density, flow and speed. Basic quantities of pedestrian dynamics are the density $\rho$ [1/m$^{2}$] in an area $A$ and the velocity $\vec{v}$ [m/s] of persons or a group of persons, and the flow through a door or across a specific line $\vec{Q} = \vec{v}(\vec{r, t}) \rho (\vec{r}, t)$ [1/s]. The measurements also yield mean values of these quantities. The task is to improve the given methods such that they allow to go fairly close to the real data of the crowd quantities. The methods presented here are based on video tracking of the head from above. Note that  tracking of e.g. a shoulder or the chest might be even better, though more difficult to obtain.
      
The density distribution knowledge in a very crowded area allows us to draw a so called density map to show us congestion directly as regions of high density.
The relationship between the pedestrian density $\rho$ and the pedestrian maximum walking speed $v_{max}$ are formalized into a graph known as the fundamental diagram $v_{max} = f(\rho)$ \cite{Fruin1971a}. Since pedestrians move slower in a region of high density, the simulated particles should update their speed with the surrounding circumstances to maximize their rate of progress towards their goals.  
 
\subsection{Data collection and type of observation}
\begin{figure}[H]
\begin{center}
\includegraphics[width=\columnwidth]{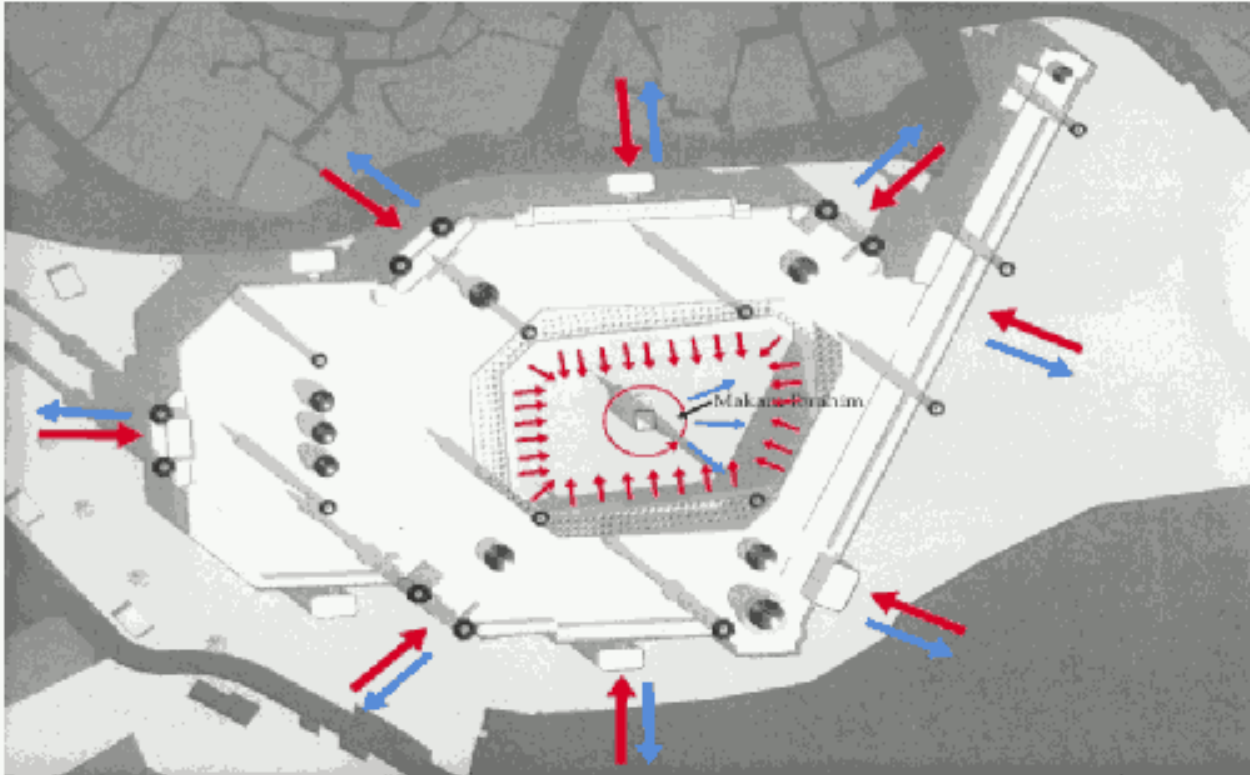}
\end{center}
\caption{Overview of the Mosque with entrance doors.}
\label{fig:haramentrances-1}
\end{figure}

Tawaf observations at the Haram mosque in Mecca were made during Hajj 2009 by Mr. Faruk Oksay. The Mataf area has 10 entrances / exits. The flow of the Tawaf is controlled. All pilgrims begin and end their Tawaf at the same place (see fig.\ref{fig:haramentrances-1}). The number of pilgrims during this period is sufficient to observe the behaviour of high density crowd dynamics.

Figure \ref{fig:haramentrances-1}. shows the main gate doors, side entrances, stairs to the Mataf open air of the Haram.

\begin{figure}[H]
\begin{center}
\includegraphics[width=\columnwidth]{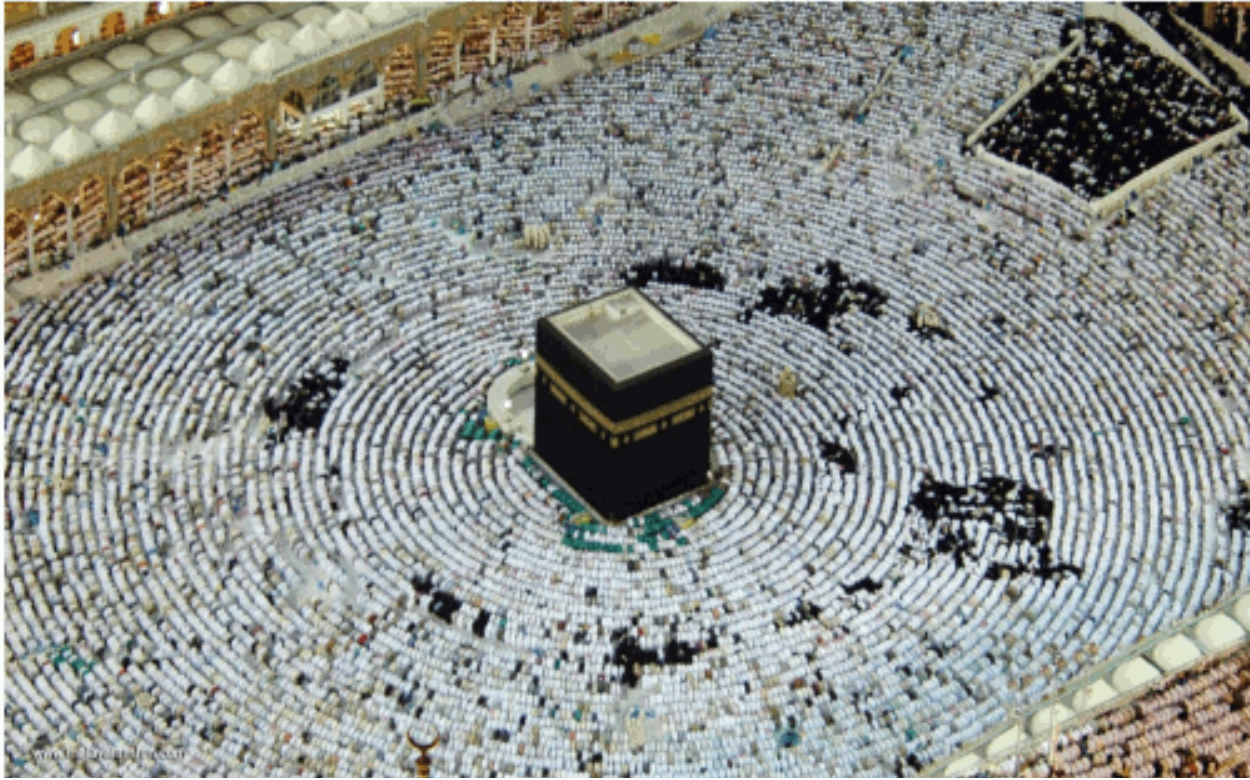}
\end{center}
\caption{Pilgrims performing the prayer ritual in the Mosque in Mecca. During performance of prayer the Tawaf come to standstill, there are no movement around the Kaaba.}
\label{fig:prayer-4}
\end{figure}

All observations took place on Friday November 27th 2009 corresponding to 10th of Dhu al-Hijjah 1430 Hijri in the afternoon.
During the total observation period of three hours, three prayers (Midday, Asr and Maghreb (sunset-prayer)) were performed, where in this time the Mataf area comes to a standstill (see fig. \ref{fig:prayer-4}). Our video observations show that the pilgrims have the desire to be near the Kaaba. Therefore approximately 70 percent (visually detected on video) of the pilgrims perform their Tawaf movement near the Kaaba wall, which causes a high density in this area.

In Figure \ref{fig:prayer-4}, one can see all of the pilgrims perform the prayer ritual in the holy mosque in Mecca.

\begin{figure}[H]
\begin{center}
\includegraphics[width=\columnwidth]{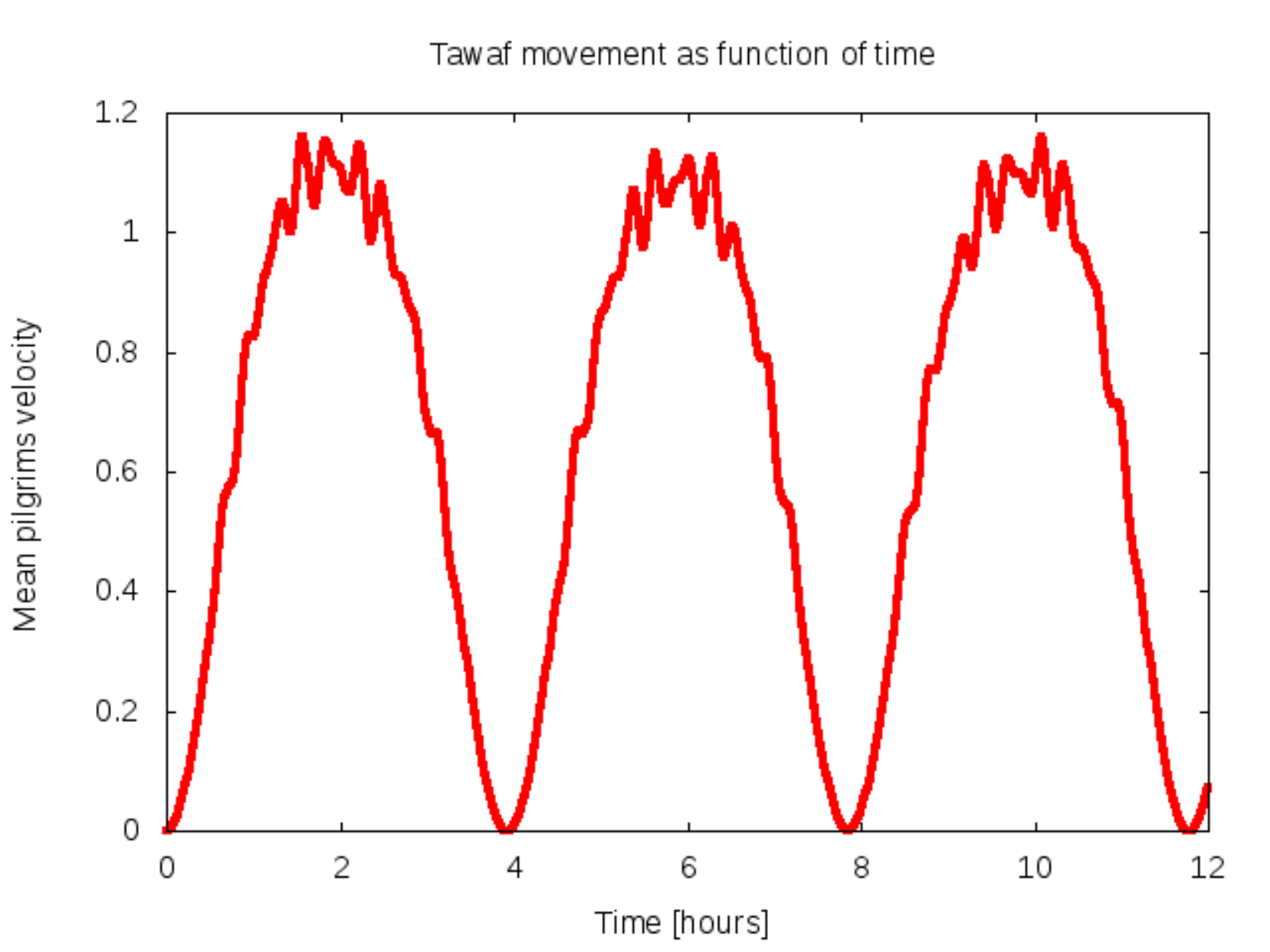}
\end{center}
\caption{Mean velocity of pilgrims in the Mataf area as a function of time over 12 hours.}
\label{fig:mataf3}
\end{figure}

The Tawaf around the Kaaba is a periodic movement for the time between two prayers. The observed number of pilgrims performing their Tawaf ritual at the Mataf area increases slowly after every prayer until the Mataf attains it's maximum capacity (see fig. \ref{fig:mataf3}).

Figure \ref{fig:mataf3} shows a typical pedestrian movement in the Mataf area over daytime. During prayer times individuals stand still and therefore movement equals approximately zero. The fluctuations in the velocity flow are created by the turbulence in the pedestrian flux. Note that the average local density in a specific location in the Mataf area exceeded 8 persons/m$^{2}$ during the Hajj periods (see fig. \ref{fig:densitydistribution-5} and \ref{fig:density-1}).
\subsection{Goals}
Our first goal is to identify new methods and create a test system capable of extracting pedestrian movement information from video, similar to that collected by our HD-Cameras in the Hajj-2009, such that any movement can be analysed to spot suspicious activity. 
This task to collect pedestrian data and extract pedestrian motion from video sequences required an involvement and development of appropriate methods, followed by further analysis of this data to identify emergent motion or crossing trajectories.

The secondary goal is to identify the limitations of the approach including the system and data requirements for the techniques to work more effectively. More specific, the project goals are: 
\begin{itemize}
\item Develop a framework for video and image analysis,
\item Develop an approach and relevant diagnostic software to collect movement data from video,
\item Identify the requirements for such methods to work effectively, such as image quality, resolution and orientation,
\item Identify how to interpret movement information,
\item Interpret the movement data and examine abnormal behaviour,
\item Design and produce a working implementation that demonstrates the above goals,
\item Identify approaches that could further improve the system.
\end{itemize}

\section{Estimation of Crowd Density}
There are different techniques developed to extract information describing the position of pedestrians in a location, but not all of them are appropriate for detecting and pursuing pedestrian movement under different and extremely weather conditions.
In their published work \cite{Papageurgiou1999}, Papageurgiou and Poggio developed a system attempting to recognize human figures based on pixel similarities through a large training set of figures under various light and weather conditions. 
To identify the movement of the figures, the system analyses the similarity between matches of consecutive frames. This method works quite well when the training set is large, but requires a high computational efficiency which achieves processing rates of 10 Hz \cite{Papageurgiou1999}. The study shows that accurate recognition can be done with coarse image data.

Another approach to estimate crowd density is based on texture analysis. Velastin et al. \cite{Marana1998} assumed that crowds with high density possess texture properties. The proposed method, texture features were computed for the whole image and applied to crowd density estimation \cite{Zhang2012}.
In particular, all displayed textures,
like wavelets \cite{Verona2001, Li2006} and the gray level dependence matrix \cite{Wu2006, Sen2009},
were used to estimate crowd density.
The results exhibit, how effective statistical analysis of texture display is compared to neural networks when measuring crowd density.
Unfortunately, this system examines only static images and cannot cover crowd motion, but the techniques can be used to track pedestrian movements.

Other strategies based on image segmentation were pursued by Heisele and Woehler \cite{heisele1998}, where raw data is filtered to split the image into segments, which are then analysed. Those images that match particular shapes are analysed further. 
This approach allows to distinguish different images with common color and luminescence.

\subsection{Data}
\begin{figure}[H]
\begin{center}
\includegraphics[width=\columnwidth]{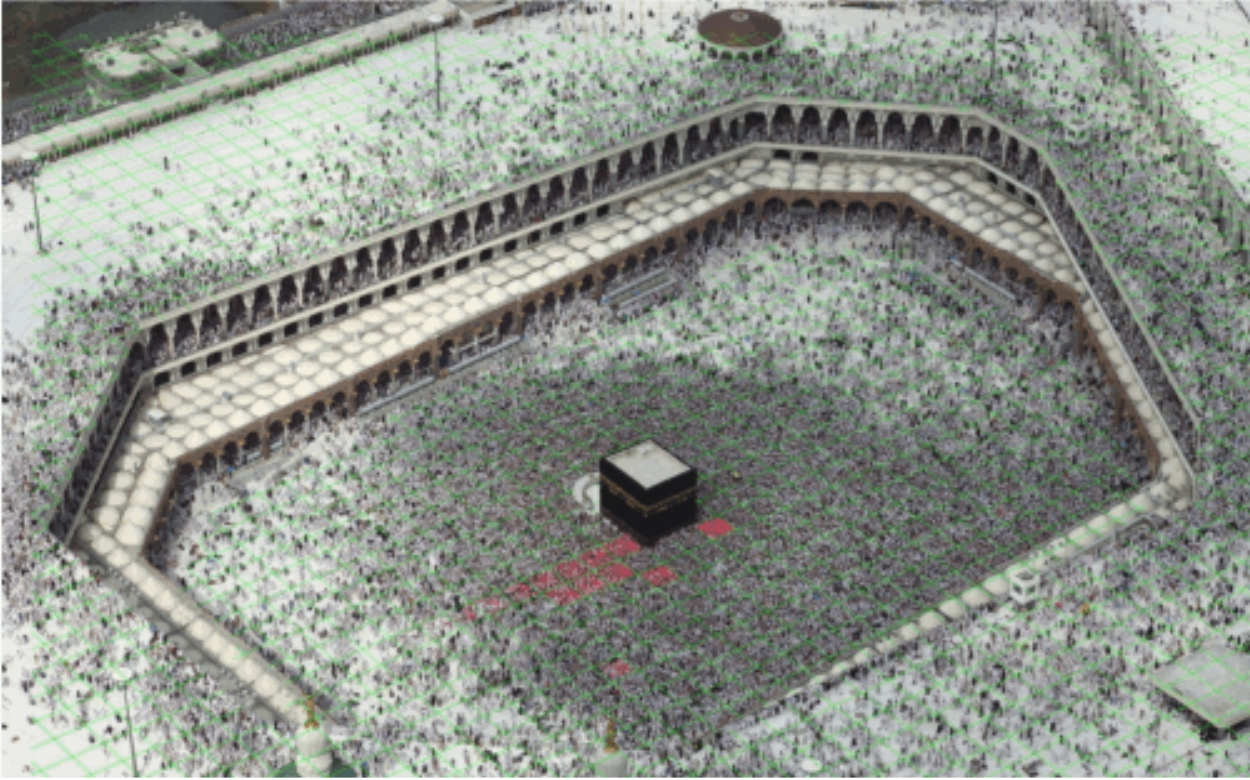}
\end{center}
\caption{The Haram piazza top view (observation area).}
\label{fig:muster-1}
\end{figure}

The required data on pedestrian behaviour (e.g. density-effect, shock-waves-effect,...) in the Haram can be done from our video recordings.  
All observed effects can be analysed by simply watching the recorded videos. But if we want to extract data like walking speeds from such observations we have to examine the videos frame by frame. This is very time consuming. As a result of this, and the need for more efficient data, the idea arose to use an automatic detection system. At that time no sufficient system was available for the detection of human bodies, therefore some essential requirements were formulated. From the requirements we derived an idea to formulate an image processing system with the help of other programs, such as Optical Flow with OpenCV (\href{http://opencv.org/}{http://opencv.org/}) and Quest3D (\href{http://www.quest3d.com/}{http://www.quest3d.com/}).
The materials used for this test are videos recorded at an
outdoor piazza of the Haram mosque in Mecca where people congregated at different times
during one day, simulating a surveillance application. The data content had a wide range of crowd densities, from very low to very high. Three different data-sets, labelled morning observation, afternoon observation and combined observation (before and after the prayer times) were used. Each data set had 20 selected images with high resolution.
Examples of images are shown in figure \ref{fig:muster-1} and \ref{fig:grid-1}.
In order to collect pedestrian data and to study pedestrian traffic flow operations on a platform in detail, observations were also made from a platform of the Haram Mosque in Mecca. These observations concerned pilgrim walking speeds and density distributions on the Mataf area and (individual) walking times as functions of the distance from the Kaaba wall.

\subsection{Methods}
\subsubsection{Manual estimation of crowd density}
\begin{figure}[H]
\begin{center}
\includegraphics[width=\columnwidth]{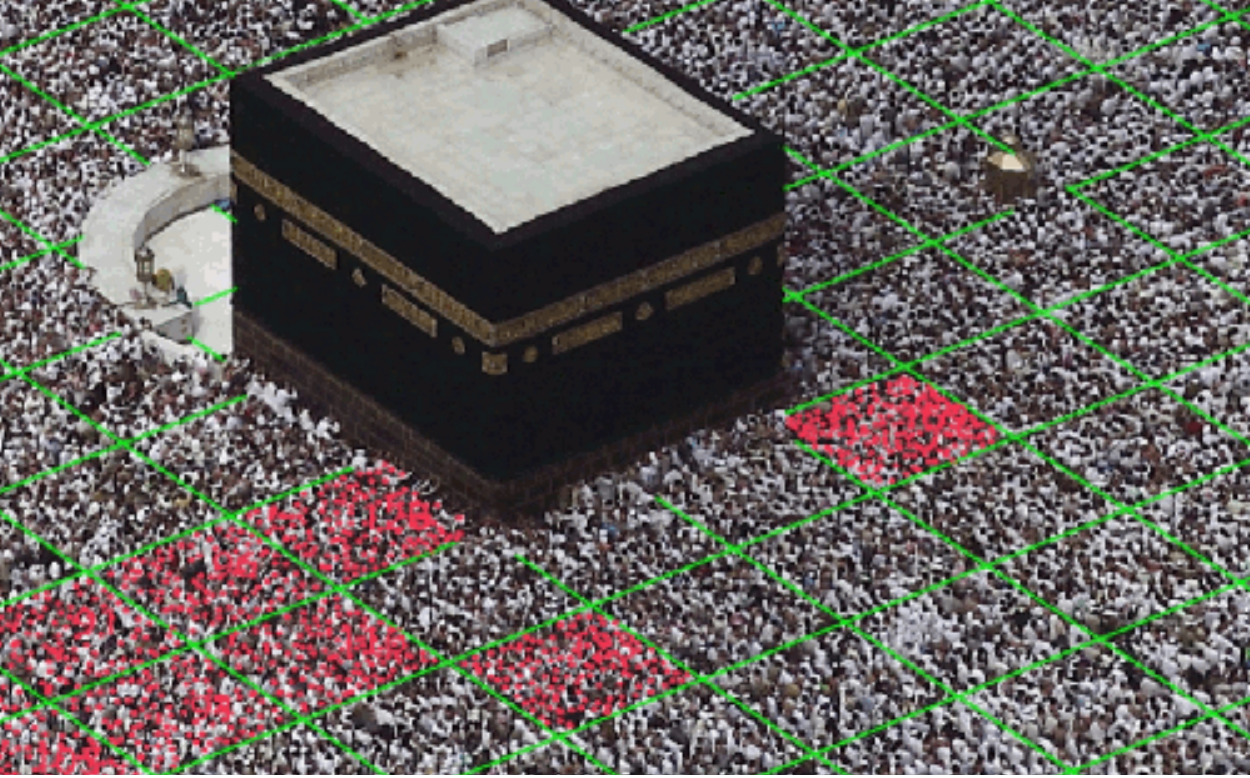}
\end{center}
\caption{Grid of regular cells with dimension of 5m$\times$5m. }
\label{fig:raster-1}
\end{figure}

\begin{figure*}[ht]
\begin{center}
\includegraphics[width=1.0\linewidth]{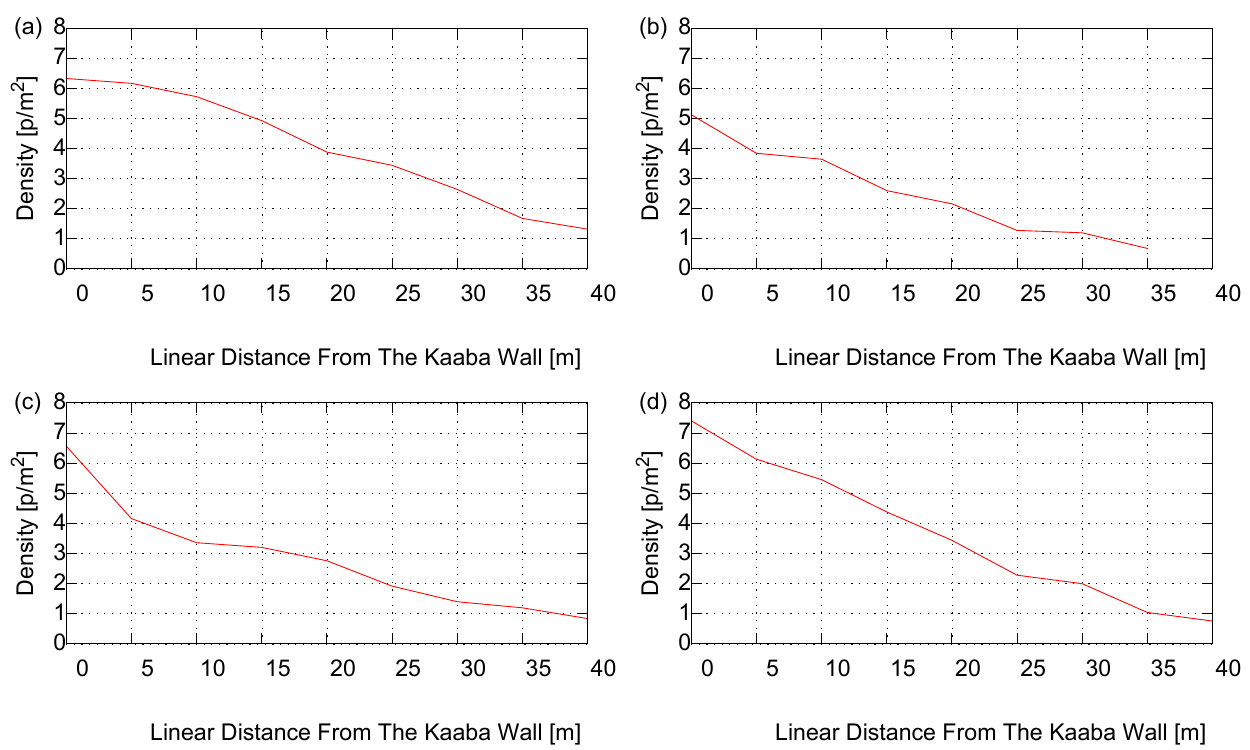}
\end{center}
\caption{Decrease of pedestrian density on the Mataf Area as function of the distance from the Kaaba wall: (a) before Mid-Day prayer; (b) shortly after Mid-Day Prayer; (c) half-hour after Mid-Day Prayer; (d) Rush Hour.}
\label{fig:densitydistribution-5}
\end{figure*}

The estimation of crowd density is an important criterion for the validation of our simulation tools. Processing is done in three levels.
\begin{itemize}
\item Existing footage is loaded on a 3D program as a backplate.
\item From several provided 2D- architectural drawings we build a 3D model of the mosque.
\item A virtual camera has to be matched in position, rotation and focal length to the original camera so that the features of the 3D-model match the features of the filmed mosque.
As the dimensions of the mosque are known, we then establish a grid of regular cells on the Mataf area, each one of which has a size of 5mx5m (see fig. \ref{fig:raster-1}).
Through image editing software, we start a manual counting process. 
This regular grid is used to observe the density behaviour over all of the Mataf area, from the nearest range to the Kaaba wall up to outside of the Mataf and the accumulation process (by the Black Stone and Maquam Ibrahim). 
The results of this investigation are shown in figures \ref{fig:densitydistribution-5} (a), (b), (c) and (d) and illustrate us the behaviour of the pilgrim density on the Mataf area at different times during the day.

\end{itemize}

With a new computer
algorithm developed within this investigation, where the Mataf area is divided in regular cells. The number of pedestrians in every cell as function of time is determined through repeating the counting process many times. The average value is identified as local density $\rho(\vec{r}, t)$. The data extracted from the videos allowed us to determine not only densities in larger areas, but also local densities, speeds and flows. As an example the density distribution on the Mataf area is shown in figure: \ref{fig:densitydistribution-5}. The data was obtained by semi-manual evaluation.
\paragraph{Dependence of the Density distribution on the Mataf as function of time}   
Figure \ref{fig:densitydistribution-5} shows density decline curves for  different distances from the Kabaa in a specific time. The curves indicate that the local density amount vary strongly over the (0 < x < 40 m) range. 

\begin{figure*}[ht]
\begin{center}
\includegraphics[width=1.0\linewidth]{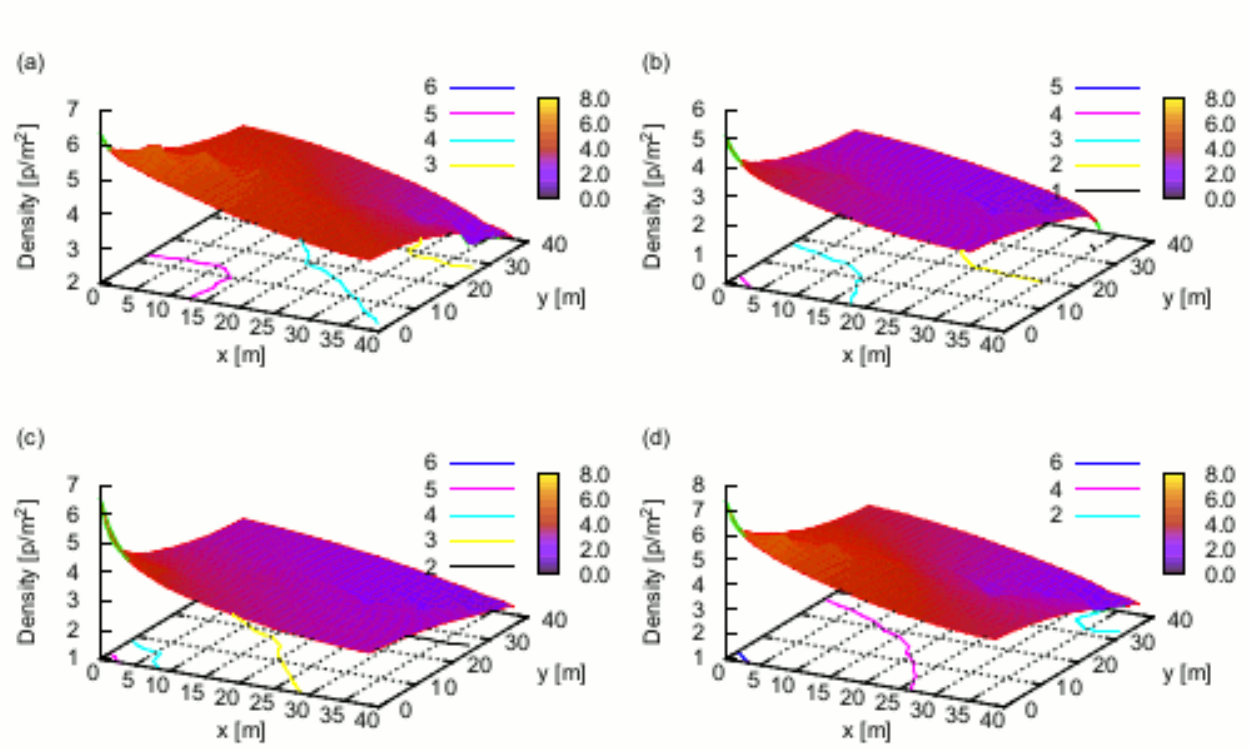}
\end{center}
\caption{The Density Distribution on the Mataf Area $\rho(\vec{r}, t)$ with $\vec{r} = (x, y)$. The figure also shows the density index (persons/m$^{2}$). (a) before Mid-Day prayer (t = t$_\textit{before Mid-Day}$); (b) shortly after Mid-Day Prayer (t = t$_\textit{shortly after Mid-Day}$); (c) half-hour after Mid-Day Prayer (t = t$_\textit{half-hour after Mid-Day}$); (d) Rush Hour (t = t$_\textit{Rush}$).}
\label{fig:density-1}
\end{figure*}

Figure:\ref{fig:density-1} shows the pedestrian density distribution on the Mataf area as a function of the position $\vec{r}$ and time $t$. One clearly recognizes density waves, with maximum density package near the Kaaba wall. There the average local density can reach a critical value of 7 to 8 persons/m$^{2}$. The congested area increases the local density to a critical and dangerous amount. As a consequence the pedestrians begin to push to increase their personal space and create shock-waves propagating through the crowd, which can be seen as density waves, or density packages. 
   
\begin{figure*}[ht]
\begin{center}
\includegraphics[width=1.0\linewidth]{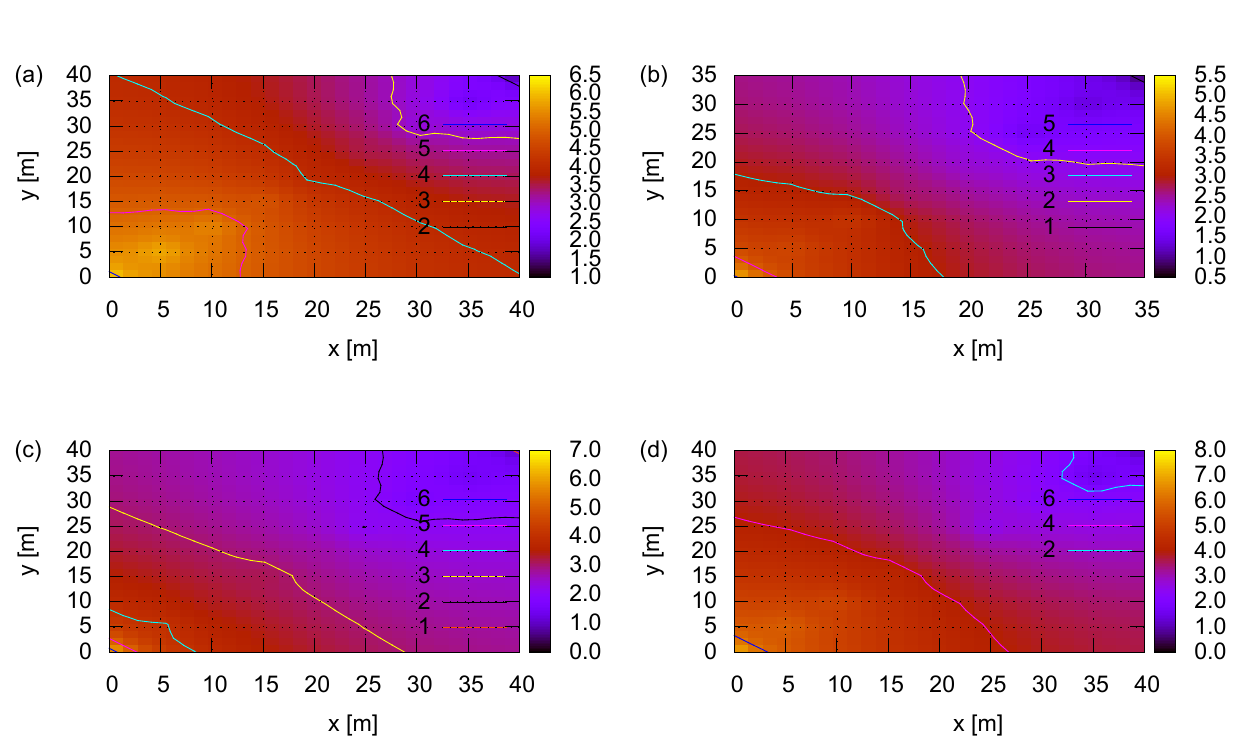}
\end{center}
\caption{The Density map indicates that the highest pedestrian density in the area of the Kaaba: (a) before Mid-Day prayer; (b) shortly after Mid-Day Prayer; (c) half-hour after Mid-Day Prayer; (d) Rush Hour.}
\label{fig:densitymap-1}
\end{figure*}

The Density map illustrates how the pedestrian density decreases from the inside to outside of the Mataf area, (see fig.\ref{fig:densitymap-1}).
As we have mentioned that in the Mataf area pedestrians move in the restricted space, the layout is gradually painted in different colors. The color of every point of the space corresponds to the current density in this particular area. The density map is constantly repainted according to the actual values: when the density changes in some point, the color changes dynamically to reflect this change. In case of zero density the area is not painted at all (see fig. \ref{fig:densitymap-1} (a), (b), (c) and (d)).

During the rush hour in a Hajj period the local density in the Mataf area reaches the maximum as we can see in the following figures \ref{fig:density-1} (d) and \ref{fig:densitymap-1} (d). The local density can reach 8 to 9 persons/m$^{2}$ in a specific time during the day. The maximal density concentrates near the Kaaba wall.

\begin{figure}[H]
\begin{center}
\includegraphics[width=\columnwidth]{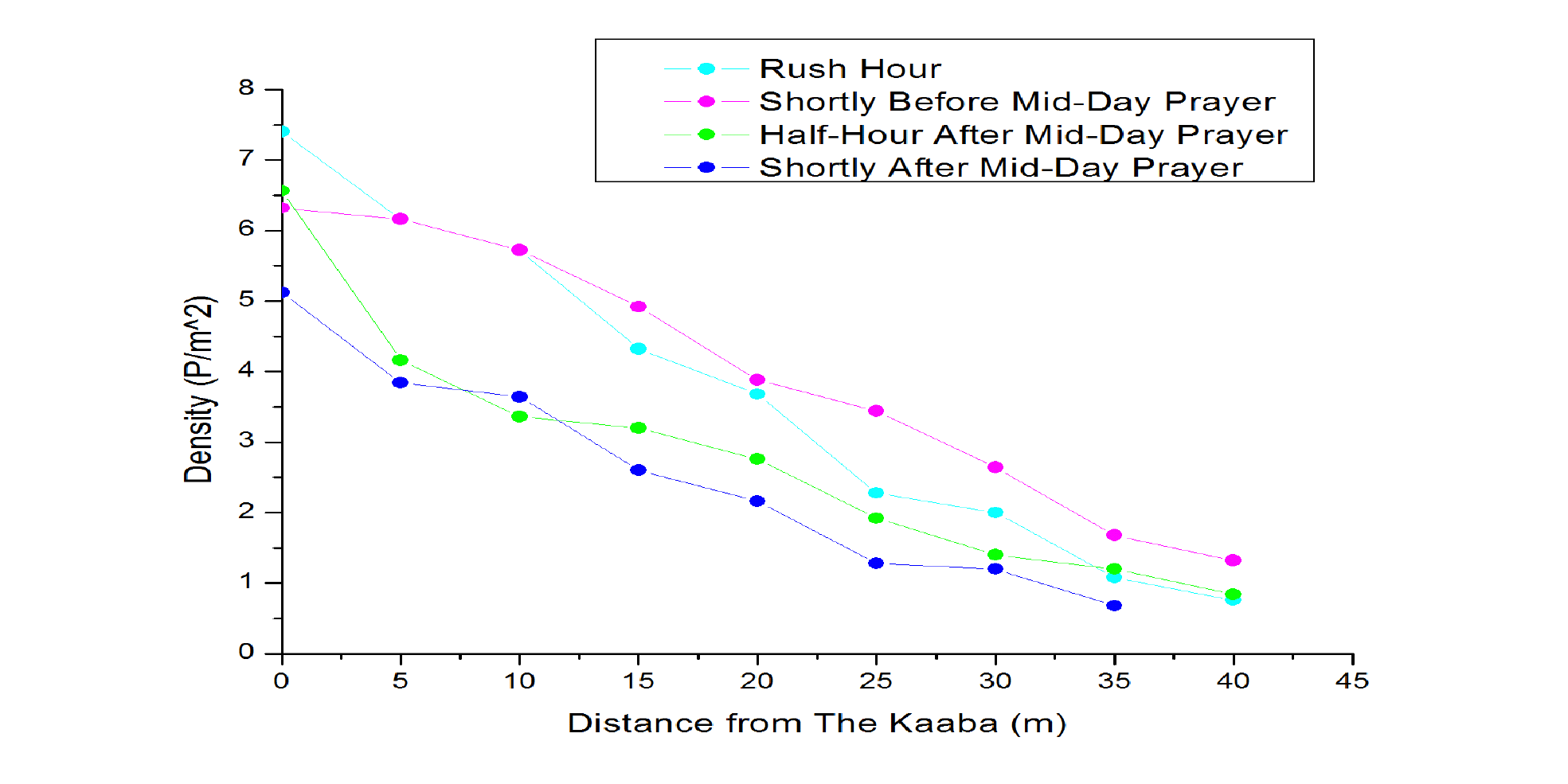}
\end{center}
\caption{Crowd density on the Mataf area in different time intervals. Highest density in the area of the Kaaba.}
\label{fig:densitydistribution-6}
\end{figure}

\paragraph{Densities over time and space}
We observe the density behaviour on the Mataf area at different times during the day, before and after the prayer, and we compare this density with the simulation density results. The maximum registered density was 7 to 8 persons/m$^{2}$ and this represents a high crowd density. 
The results of the estimation based on the statistical method, presented in figures \ref {fig:densitydistribution-5},\ref{fig:density-1} and \ref{fig:densitymap-1}, reached a mean of 92 percent correct estimations. It is possible to verify that the results were quite good for all evaluated images except for the one made up of high density crowd images, which reached only 84 percent correct estimations. In the Mataf area, near the black stone, the pilgrim density reached over 9 persons/m$^{2}$. For this reason it is very difficult to recognize and track every head and as a result, a 100 percent correct estimation would be very difficult. All statistical results illustrating the density distribution at the Mataf area at different time intervals are demonstrated in the figure \ref{fig:densitydistribution-6}.   

\subsubsection{Automatic estimation of crowd density}
\begin{figure}[H]
\begin{center}
\includegraphics[width=\columnwidth]{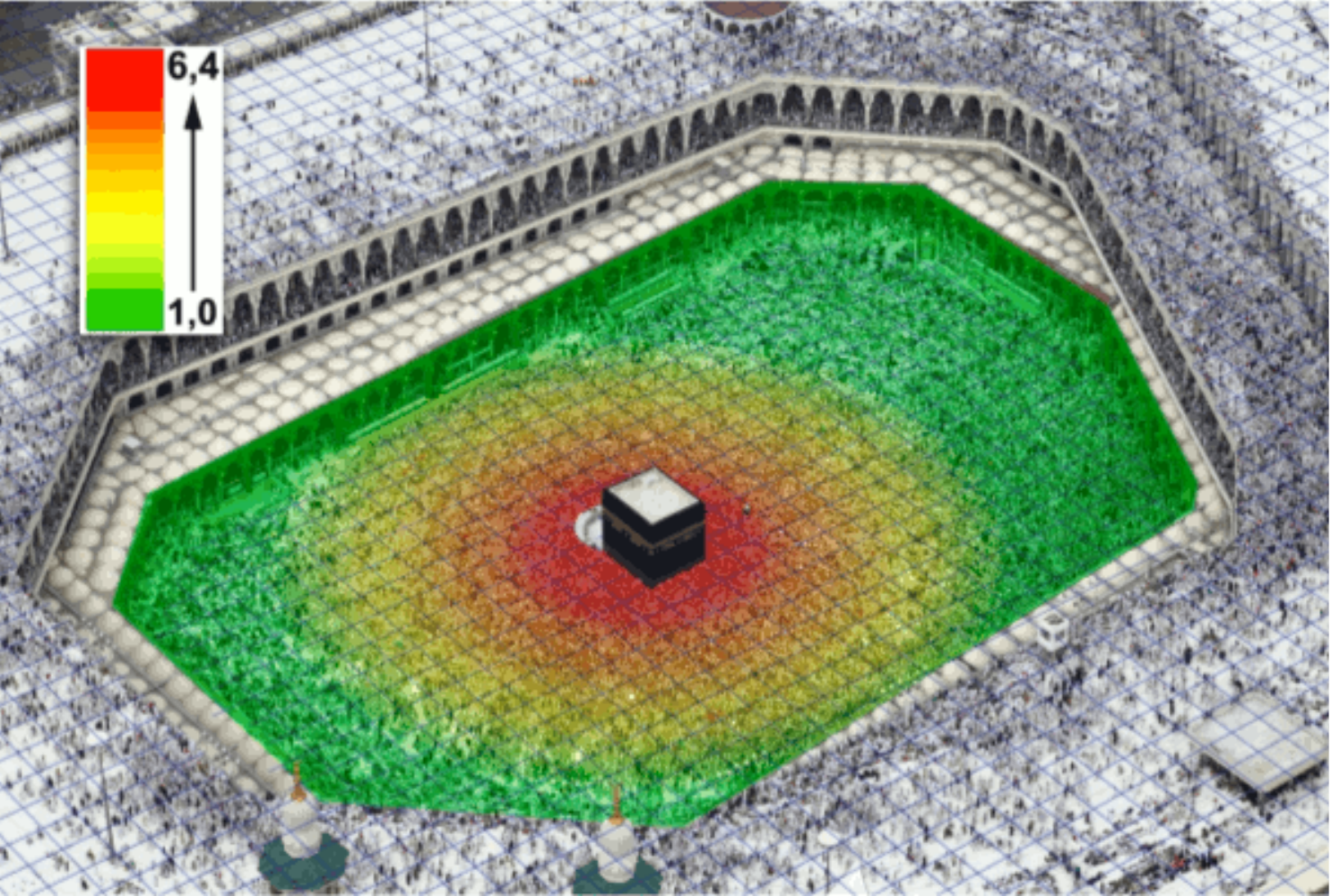}
\end{center}
\caption{Density distribution in the Mataf area before Mid-Day prayer. Red color indicates high density, where green color indicates low pilgrims density.}
\label{fig:colordensitydistribution-1}
\end{figure}

This part of the dissertation considers the role of automatic estimations of crowd density and their importance for the automatic monitoring of areas where crowds are expected to be present. A new technique is proposed which is able to estimate densities ranging from very low to very high concentrations of people. This technique is based on the differences of texture muster on the images of crowds. Images of low density crowds exhibits rough textures, while images with high densities tend to present finer textures. The image pixels are classified in different texture classes, and statistics of such classes are used to estimate the number of people. The texture classification and the crowd density estimation are based on self-organizing neural networks. Results obtained estimating the number of people in a specific area of the Haram Mosque in Mecca are presented in figure \ref{fig:colordensitydistribution-1}).

\subsection{Data Analysis}
In the latter paragraphs we focus on crowd density estimation for several reasons. According to the crowd disasters study by  Helbing and Johansson \cite{Helbing2007}, one of the most important aspects to keep a crowd safe is to predict and identify areas with high density crowds preventing large crowd pressures to be built up. Areas where crowds are likely to build up should be identified prior to the event or
operation of the venue. This is important as crowds usually
exist in certain areas or at particular times of the day. Places
where crowd density rises up over time are likely to congest and need careful
observations to ensure the crowd safety. Basically, crowd density surveillance and estimation can be a good solution for management and controlling the crowds safety.

The results of the estimations obtained during the tests allow us to consider both methods successfully. While the statistical method reached quite good estimation rates (around 92 percent) for most groups, the spectral method illustrated small deviations between the best and the worst estimations, reaching on average almost the same rates of correct estimation obtained by the statistical method.

\section{Method of getting the pedestrian speed}
As speeds are hard to observe, walking times were measured, from which walking speeds were derived. 
In addition to walking times and pedestrian densities other variables needed to be considered to complete the input of the simulation model (such as the number of in and out going pilgrims and the configuration of the structure during the rush hour at the Hajj period).
The observables are the walking time, velocities and the corresponding densities of the pilgrims performing their Tawaf and Sa'y.
The movements of the pilgrims going in and out of the Haram give us data to calculate the flux related to the Tawaf. The distribution of both in and out going pilgrims over the Haram can be derived from this data.
The second type of observation concerns individual walking times. In order to measure the pilgrims' walking times in and out of the Haram, pilgrims were recorded from the moment they started walking from one spot to another, either on the piazza or going up the stairs.
The start and duration of activities, such as Tawaf or Sa'y, were measured also. Finally, locations of origin, destination and possible activities of the pilgrims were registered. To do this, the piazza is divided into small areas with a length of 5$\times$5 meters. We also recorded the movements of the pilgrims at specific moments, such as prayer times when the number of pedestrians increases dramatically. Therefore, cumulative flow curves can be constructed, out of which densities can be derived. These curves can be compared with the reference curves of Predtetschenski-Milinski \cite{Predtetschenski-Milinski1969}. 

\subsection{Subject Selection}
Data was collected on a specific subject group of pedestrians who appeared
to be 40 years of age or older. On the roof of the Mataf area we selected our tracking subjects, consisting of adult men, women and people in wheelchairs. The following individuals
were specifically not considered:
\begin{itemize}
\item Children under 13 years of age,
\item Pedestrians carrying children, heavy bags, or suitcases,
\item Pedestrians holding hands or assisting others across the Mataf,
\item Pedestrians using a quad pod cane, walker, two canes, or crutches.
\end{itemize}

To accurately quantify the normal walking speeds of the various
subject groups, pedestrians who exhibited any of the following
behaviour were also not considered:
\begin{itemize}
\item Crossing of the Mataf path diagonally,
\item Stopping or resting in the Mataf area,
\item Entering the roadway running (anything faster than a fast walk),
\end{itemize}

The pedestrian sex (male or female) of each individual in the Mataf area was recorded, as well as whether he or she was walking alone or in a group. 
The group size was also
noted when applicable. A group was defined by two or more pilgrims
walking the Mataf trajectory at about the same time, regardless of
whether or not they were apparently friends or associates. In the Mataf area, the pedestrian groups can reach 30 pilgrims walking together in the pedestrian stream.
In addition, subjects paths were monitored to determine when
they started and ended their Tawaf. Being
inside the Mataf was defined as being within or on the
painted Tawaf walking lines.
Other pedestrian behaviour was recorded when if occurred:
\begin{itemize}
\item Confusion (hesitation, sudden change in direction of travel or change of point of interest) exhibited before walking,
\item Confusion exhibited after entering the Mataf trajectory,
\item Cane use,
\item Following the lead of other pedestrians,
\item Stopping in the walking path during the Tawaf movement,
\item Difficulty going into Mataf,
\item Difficulty going out of the Mataf.
\end{itemize}
Several methods were developed 
to check the accuracy and performance of walking speed estimation abilities of the observers.
First, the walking speed was measured at the same time by three observers,
then correlations between the estimates of all observers were
determined. 
In particular, the walking velocity of one pilgrim was measured by the three observers and the mean value was taken. The results of these verification procedures are discussed after the next section. 

\subsection{Manual methods}
\begin{figure}[H]
\begin{center}
\includegraphics[width=\columnwidth]{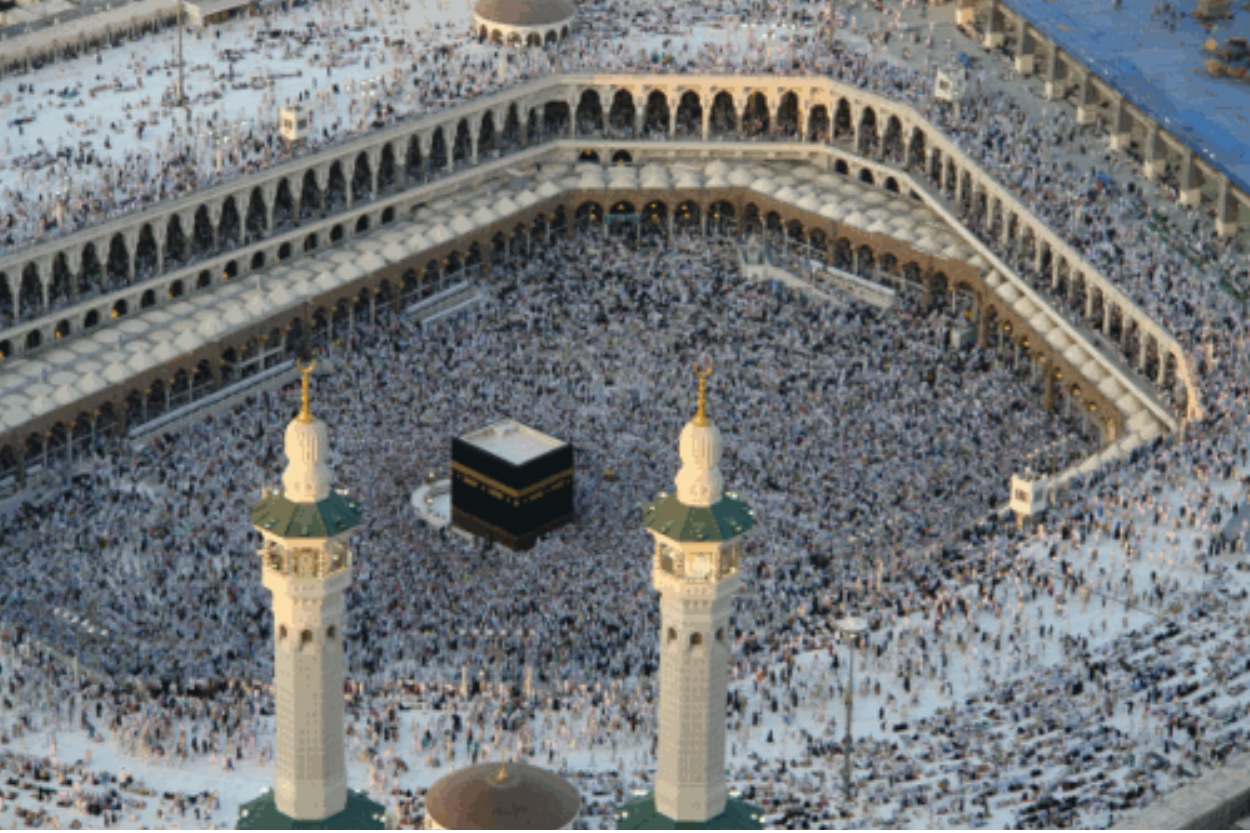}
\end{center}
\caption{Overview of the piazza of the Haram, the place where our observations are made. With a digital clock the individual walking times $t_{p}$ are measured. Since the distance between the two minarets is known from the architectural plan of the Haram, the average of the local pedestrian velocities $v(\vec{r}, t) = \parallel \vec{v}(\vec{r}, t) \parallel$ can be determined.}
\label{fig:minaret-1}
\end{figure}

\begin{figure}[H]
\begin{center}
\includegraphics[width=\columnwidth]{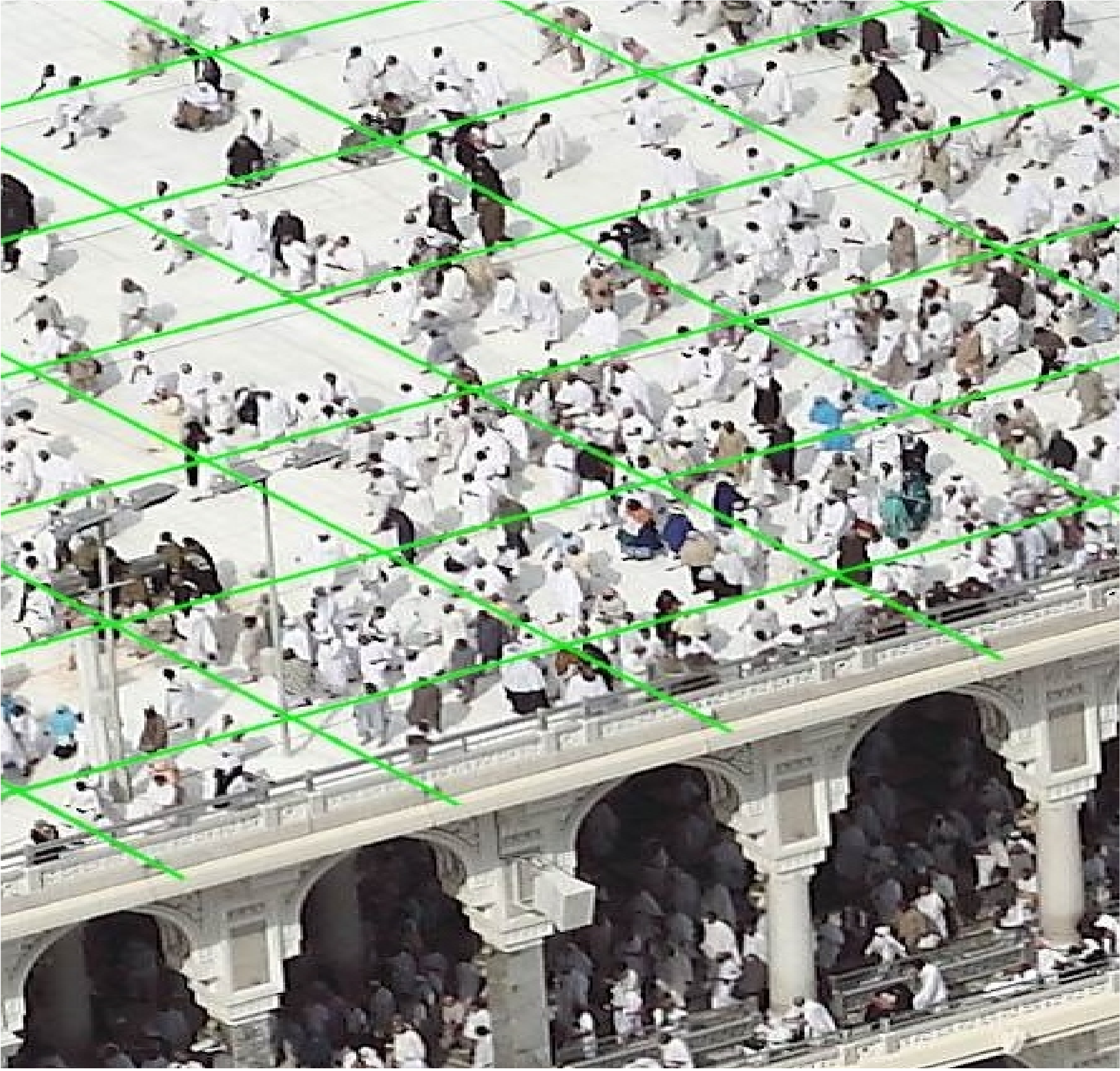}
\end{center}
\caption{Grid of regular cells with dimension of 5m$\times$5m. With the help of the regular cells and the distance between two minarets in the Haram  (see fig. \ref{fig:minaret-1}) the (individual) walking times $t_{p}$ are determined and the average of the local speeds $v(\vec{r}, t) = \parallel \vec{v}(\vec{r}, t) \parallel$ is calculated. The average walking speed for male pedestrians is 1.37 m/s, female 1.22 m/s and for people moving on wheelchairs 1.534 m/s.}
\label{fig:grid-1}
\end{figure}

From our video recordings we choose places between two minarets as references, (see fig. \ref{fig:minaret-1}). As the dimensions of the mosque were known, we then established a grid of regular cells covering all of the Mataf area, each one having a size of 5mx5m (see fig. \ref{fig:grid-1}). The distance between the two minarets is known.   
Pedestrian crossing times were measured with a digital timer and an
electronic stopwatch was implemented and synchronized with the timer of the video recorder. The watch was started as the subject stepped
off the first minaret and stopped when the subject stepped out on the opposite
minaret after crossing all the distance between the two minarets.

\subsubsection{Verification of Observer Walk-Speed Estimates and
Start-up Time Measurement}
From the roof of the Mosque every pedestrian can be identified. To establish the ability of the field  observers to identify the fitness level or the age of pedestrians with high accuracy a simple verification procedure was performed. 
The age estimation and the level of fitness of the pedestrians was based on their walking speed. It is a physio-medical fact that older pedestrians walk more slowly than younger ones (this is easily supported by field data), however, the published or already existing data on walking speeds and start-up times (i.e. the time from the beginning of a Tawaf movement until the pedestrian steps off the Mataf) have many shortcomings. Here we consider the complicated movement of the Tawaf and the human error rate of the observer. The walking speed on the Mataf area can be affected by many factors, one of the relevant factors is the age of the pedestrian. This demonstrates that the observations were quite good at identifying older pedestrians or pedestrians with fitness deficiency or physical health problems.
A digital stopwatch was integrated with the video recording sophisticated for the
measurements of pedestrian crossing times. The crossing
times of the same pilgrims were measured during five rounds of the Tawaf and the average value was determined.

\subsubsection{Pedestrian Walking Speeds Results}
\begin{figure*}[ht]
\begin{center}
\includegraphics[width=1.0\linewidth]{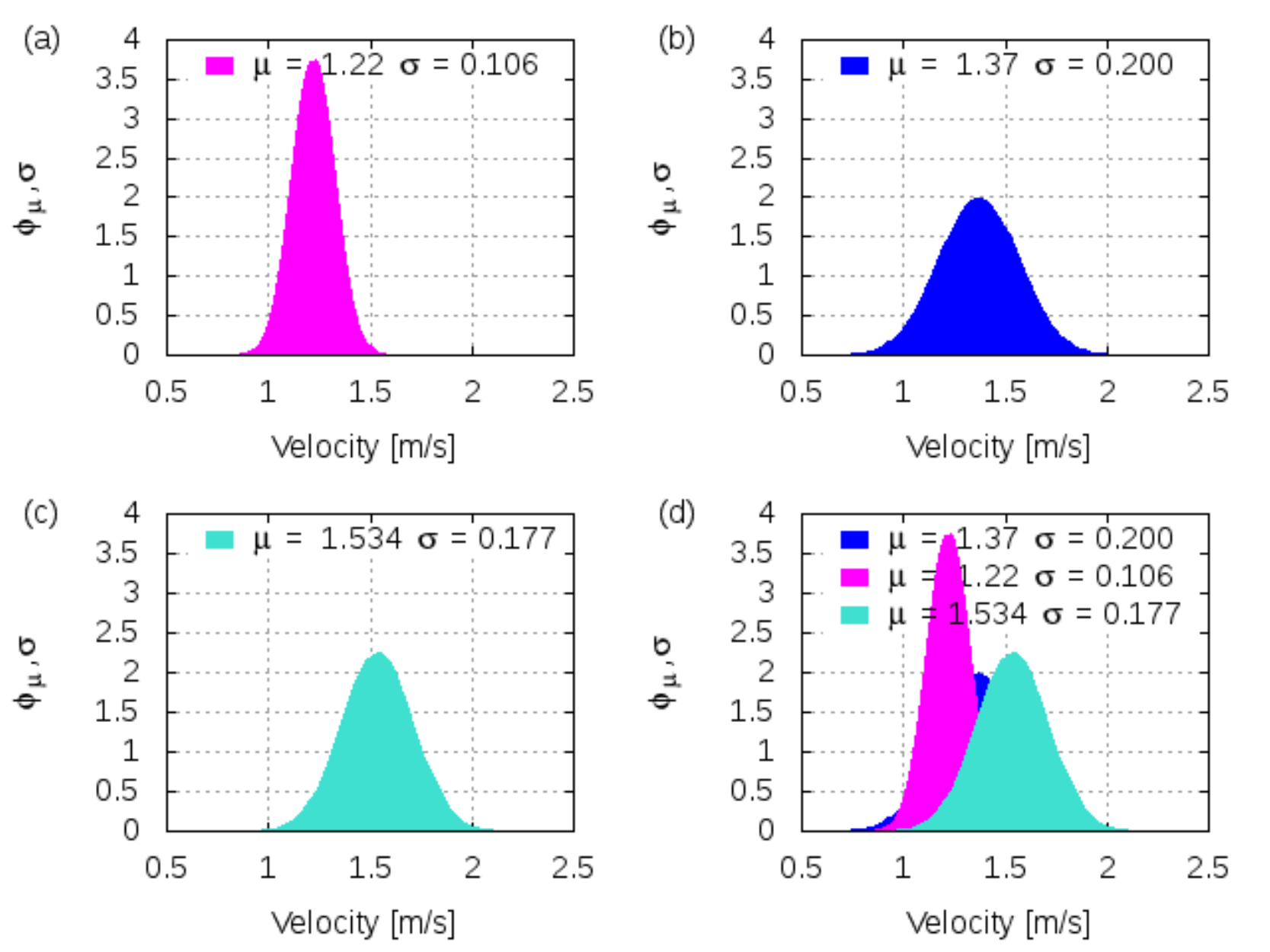}
\end{center}
\caption{Walking speed distribution in Tawaf movement. (a) Women, (b) men, (c) wheelchairs and (d) shows a comparison of the three distributions. The average walking speed for female pedestrians is $\mu = 1.22$ m/s with a standard deviation of $\sigma = 0.106$ while for male pedestrians $\mu = 1.37$ m/s and $\sigma = 0.200$ and for pilgrims on wheelchairs $\mu = 1.534$ m/s and $\sigma = 0.177$.}
\label{fig:velocitydismen-1}
\end{figure*}

 This research also examined the impact of the building layout on the pedestrian speed distribution and the pedestrian density of pilgrims performing the Tawaf movement around the Kaaba. The set of data of pedestrian walking speeds which were obtained through analysing video recording using a set of statistical techniques are displayed in figures \ref{fig:velocitydismen-1} (a), (b) and (c). The results revealed that walking speed seems to be following a normal distribution no matter of male, female, older or younger. The average speed of young people is dramatically larger than that of older people, and the average speed of male is slightly larger than that of female. The width of the obtained curves is related to the different standard deviations.
 
The mean computed walking speed represents the speed that 
85 percent of pedestrians did exceed. A total of 250 pedestrians
were observed. Included were 100 male pedestrians of about 60 years of age, 100 women pedestrians and 50 wheelchair pedestrians. This data describes all of the pedestrians observed: those walking in the center of the stream and those walking by the edge of the Mataf trajectory. As is subsequently described,
those who were walking by the edge of the Mataf tended to walk more quickly. All observed pedestrians moved in a rotational motion around the Kaaba counter-clockwise (Tawaf), in compliance with the pilgrim stream.

The mean walking speed for male pedestrians was 1.37 m/s and 1.22 m/s for female pedestrians. In conjunction with pilgrims old, the mean walking speed for younger pedestrians was 1.48 m/s
and 1.20 m/s for older male pedestrians. The results revealed that the average walking speed for young women are 1.32 m/s and 1.12 m/s for old women.   
This means
\begin{itemize}
\item Young male pedestrians had the fastest mean walking speeds [1.48 m/s] and older females had the slowest [1.12 m/s]. The differences between young men and young women [0.16 m/s] and between older men and older women [0.1 m/s], this result shows a little deviation that can be traced back to the fitness level of pedestrian or other factors, in the normal condition are approximately the same.
The mean walking speed for the
younger pedestrians ranged from 1.37 to 1.57 m/s across all conditions, with an overall mean speed of 1.48 m/s. The means for the older pedestrians range from 0.97 m/s to 1.26 m/s, with an overall mean speed of 1.18 m/s. For design purposes a mean speed of 1.33 m/s appeared appropriate;
 
\item Locations by the edge of the Mataf had faster walking speeds
because such locations has a lower pedestrian density. It is clear that the pedestrians near the Kaaba had a short walk path but in this places densities of 7 to 8 persons/ m$^{2}$ can be exceeded, making the movement of pilgrims very slow and turbulent;

\item Places situated further away from the Kaaba wall also tended to be associated with faster walking speeds. It is known from other fundamental diagrams, that pedestrians tend to walk faster along a free walkway. As might be expected
the walking speeds associated with various factors. The motion of a single individual at any given time and the direction and speed result in a long list of possible (and very likely conflicting) forces and circumstances. 
\end{itemize}

The data taken show that each of the locations and surrounding factors have a significant effect on the behaviour and walking speed of the pilgrims on the Mataf area, not forgetting that the age of the pedestrians play a significant role on the Tawaf movement and density peaks and jams are caused by pilgrims of age 70 and more. For approximately one half of the location, the factors examined there also showed an important correlation between pedestrian age, the location and the mean walking speed of the pilgrims. This funding is consistent with results published by Knoblauch  \cite{Knoblauch1996}.  

The walking speed of pilgrims shows statistically significant variations across a variety of sites, times and environmental conditions (pedestrian density on the Mataf area). On the roof of the Mosque the pilgrim density is low and every pedestrian can walk with his desired velocity. However, the mean walking speed data is explicit by clustered for both pedestrians sex, men and women, independent of the age of the pilgrims are considered.

\subsection{Automatic Estimation of Pedestrian Walking Speeds}
There exist numerous methods that track the movement of single individuals by inspecting their orientation and limb positions. 

This section highlights a real-time system for pedestrian tracking from sequences of high resolution images acquired by a stationary (high definition) camera. The objective was to estimate pedestrian velocities as a function of the local density.  
With this system the spatio-temporal coordinates of each pedestrian
during the Tawaf ritual were established. 
Processing was done through the following steps:
\begin{itemize}
\item Existing footage was loaded onto a 3D program as a backplate.
\item From several provided 2D- architectural drawings, a 3D model of the mosque was built.
\item A virtual camera was matched in position, rotation and focal length to the original camera so that the features of the 3D-model matched the features positioned on the filmed mosque.
\item Individual features were identified by eye, contrast is the criterion
\item We do know that the pilgrims walk on a plane, and after matching the camera we also obtained the height of the plane in 3D-space from our 3D model.
\item A point object was placed at the position of a selected pedestrian. During the
animation we set multiple animation-keys (approx every 25 to 50 frames (equals 1 to 2 seconds)) for the position, so that the position of the point and the pedestrian overlay nearly all the time.
\item By evolving the point with time we obtained the distance travelled, by
measuring the distance from frame to frame. We also knew the time
elapsed from the speed per frame, and hence the speed could be calculated. 
\end{itemize}

\begin{figure}[H]
\begin{center}
\includegraphics[width=\columnwidth]{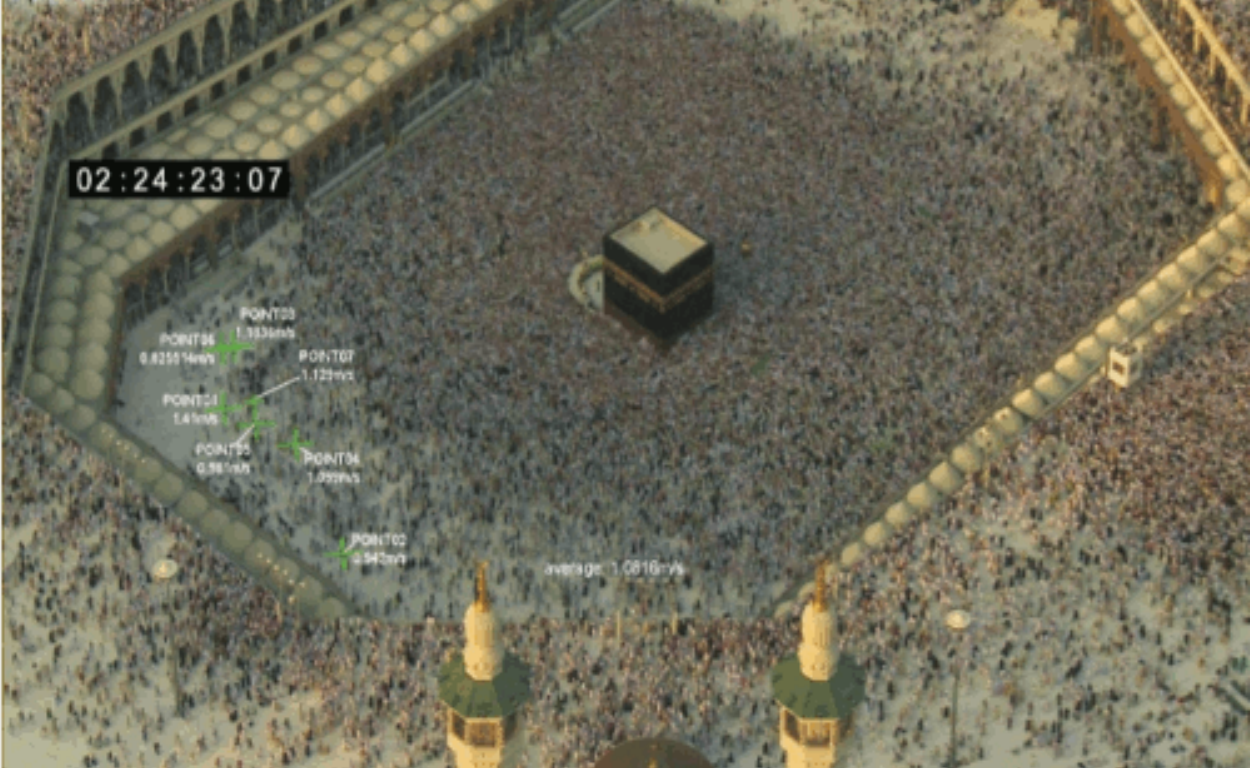}
\end{center}
\caption{Pilgrims' walking speeds on the edge of the Mataf. The average walking speed is 1.0816 m/s.}
\label{fig:walkingspeed-1}
\end{figure}

\begin{figure}[H]
\begin{center}
\includegraphics[width=\columnwidth]{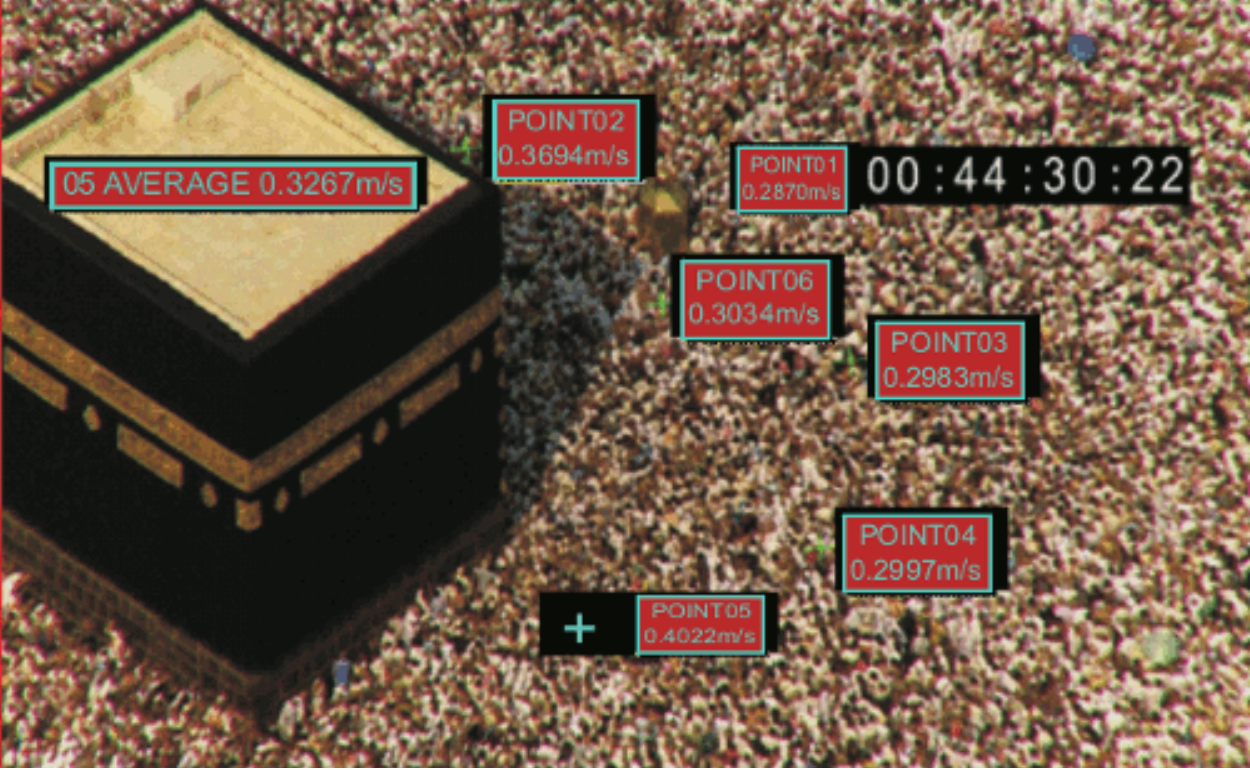}
\end{center}
\caption{Pilgrims' walking speeds in the Mataf area (near the Kaaba wall). The average walking speed is 0.3267 m/s.}
\label{fig:walkingspeed-3}
\end{figure} 

\subsection{Analysis}
From Figures \ref{fig:walkingspeed-1} and \ref{fig:walkingspeed-3} we see that the edge of the Mataf moves faster than the center, this phenomenon being known as the Edge Effect. The Edge Effect occurs when the edges of a crowd move faster than the center of the crowd. The density becomes higher and higher as one moves from the edge of the Mataf towards the center. This phenomenon is explained by the fact that all pilgrims want to be near the Kaaba wall. As a result, we find the density near the Kaaba to be the maximum density. This data can be used in validating of simulation tools.
The mean walking speed for a group of pedestrians moving in the pilgrim stream around the Kaaba was 1.0816 m/s at the edge of the Mataf and it was 0.3267 m/s for the same pedestrians groups moving inside the Mataf. These findings agree well with the statistical results discussed in a previous section.  

\section{Comparison of walking speeds}
\begin{figure}[H]
\begin{center}
\includegraphics[width=\columnwidth]{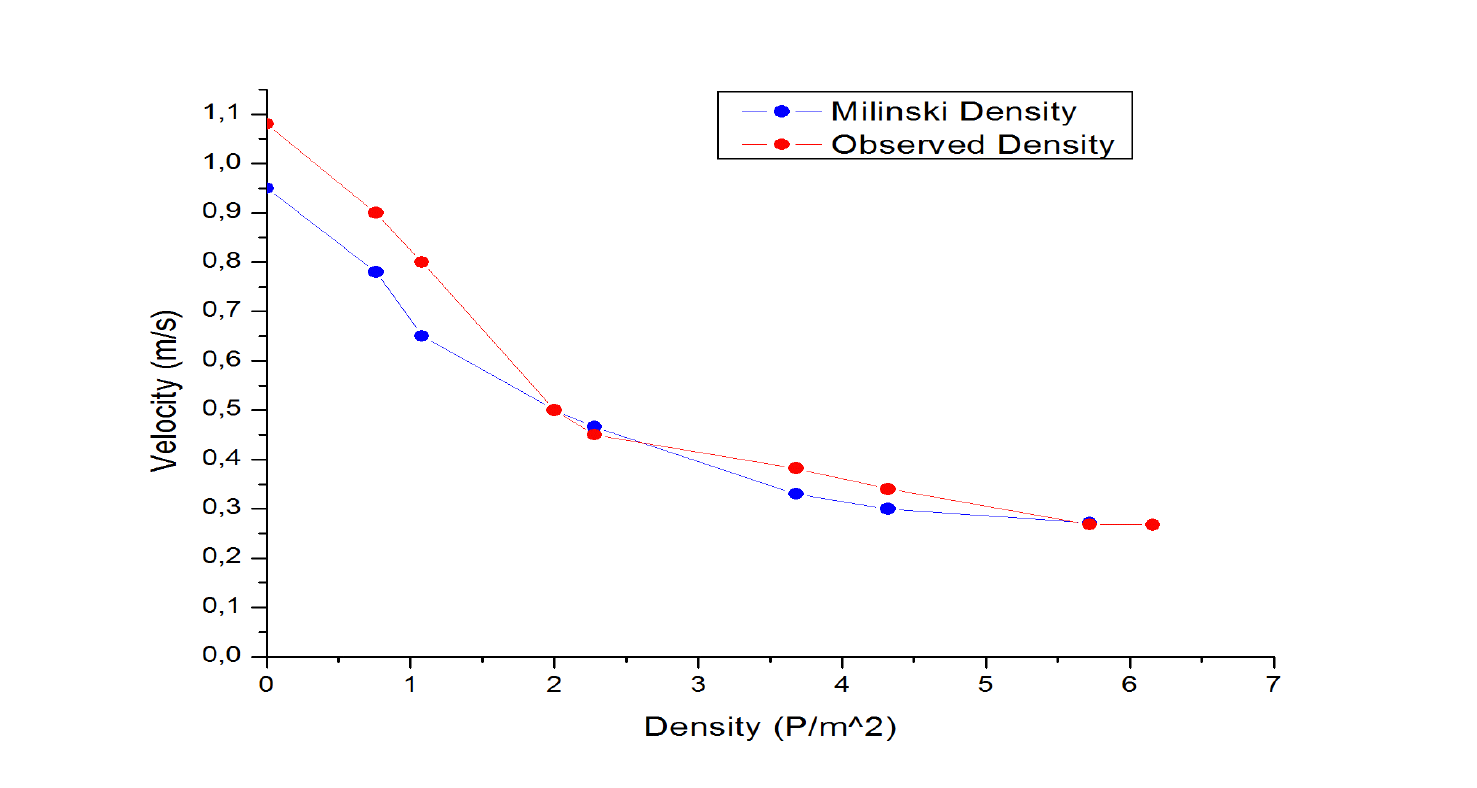}
\end{center}
\caption{The Velocity-Density-Diagrams half-hour after Mid-Day Prayer (t = t$_\textit{half-hour after Mid-Day}$). Average of the local speeds $\vec{v}(\vec{r}, t)$ as a function of the local density $\rho(\vec{r}, t)$. Our
own data are shown as red points.
The blue points correspond to the data obtained by (PM) \cite{Predtetschenski-Milinski1969}. The difference in velocity at lower densities can be explained by the fitness level of pedestrians.}
\label{fig:comparison-1}
\end{figure}

One of the must-have results is to compare the mean values and variances of walking speeds in both observations and simulation results. A distinction will be made for walking speeds inside and outside of the Mataf platforms.
We made a comparison between our plots derived from the video observation and the fundamental diagrams of (cf.\ fig. \ref{fig:comparison-1}):
\begin{itemize}
\item Walking speeds:
\begin{itemize}
\item On the edge of the Mataf (free flow speed) where the pedestrian density is lower than 3 persons/m$^{2}$.
\item On the center of the Mataf.
\item On the Mataf inside near the Kaaba wall where the pedestrian density attains extreme levels (8-9 persons/m$^{2}$).
\end{itemize}
\end{itemize}

All well-known fundamental diagrams predict the same behaviour and have the same properties: speed decreases with increasing density. So the discussion above indicates there are many possible reasons and causes for the speed reduction. For example there is a linear relationship between speed and the inverse of the density for pedestrians moving in a straight way \cite{Seyfried2005}. However the pedestrian walking speed can be affected by internal and external factors (such as the amount of pedestrian inflow and outflow as well as the configuration of the infrastructure) not to forget the physiology of the human body. It is found that individuals walk faster in outdoor facilities than in corridors \cite{Lam2000}. According to Predtechenskii and Milinskii (PM) the average walking speed depends on the the walking facility \cite{Predtetschenski-Milinski1969}.  
In other circumstances Weidmann confirmed a linear relationship between the step size length of walking pedestrians and the inverse of the density \cite{Weidmann1993}. The small step size means low pedestrian velocity, caused by reduction of the available space with increasing density.
The discussion above shows that there are many possible factors influencing the fundamental diagram. To identify these factors, it is necessary to exclude as many influences of measurement methodology and short range fluctuations from the data.
Figure \ref{fig:comparison-1} shows the average local speed $\vec{v} (\vec{r}, t)$  as a function of the local density $\rho(\vec{r}, t)$ half-hour after Mid-Day Prayer (t = t$_\textit{half-hour after Mid-Day}$). Our own data is shown as red points. The blue points correspond to the Milinski fundamental diagram.
Moreover investigation data analysing the Mataf area represented by blue points in figure \ref{fig:comparison-1} and showed that a reduction of the available navigation space illustrates the causes responsible for the speed reduction with density in pedestrian movement. The small deviation in pedestrian walking speed at lower density can be explained by the fitness level of the pedestrian.

\section{Movement Recognition}
\begin{figure}[H]
\begin{center}
\includegraphics[width=\columnwidth]{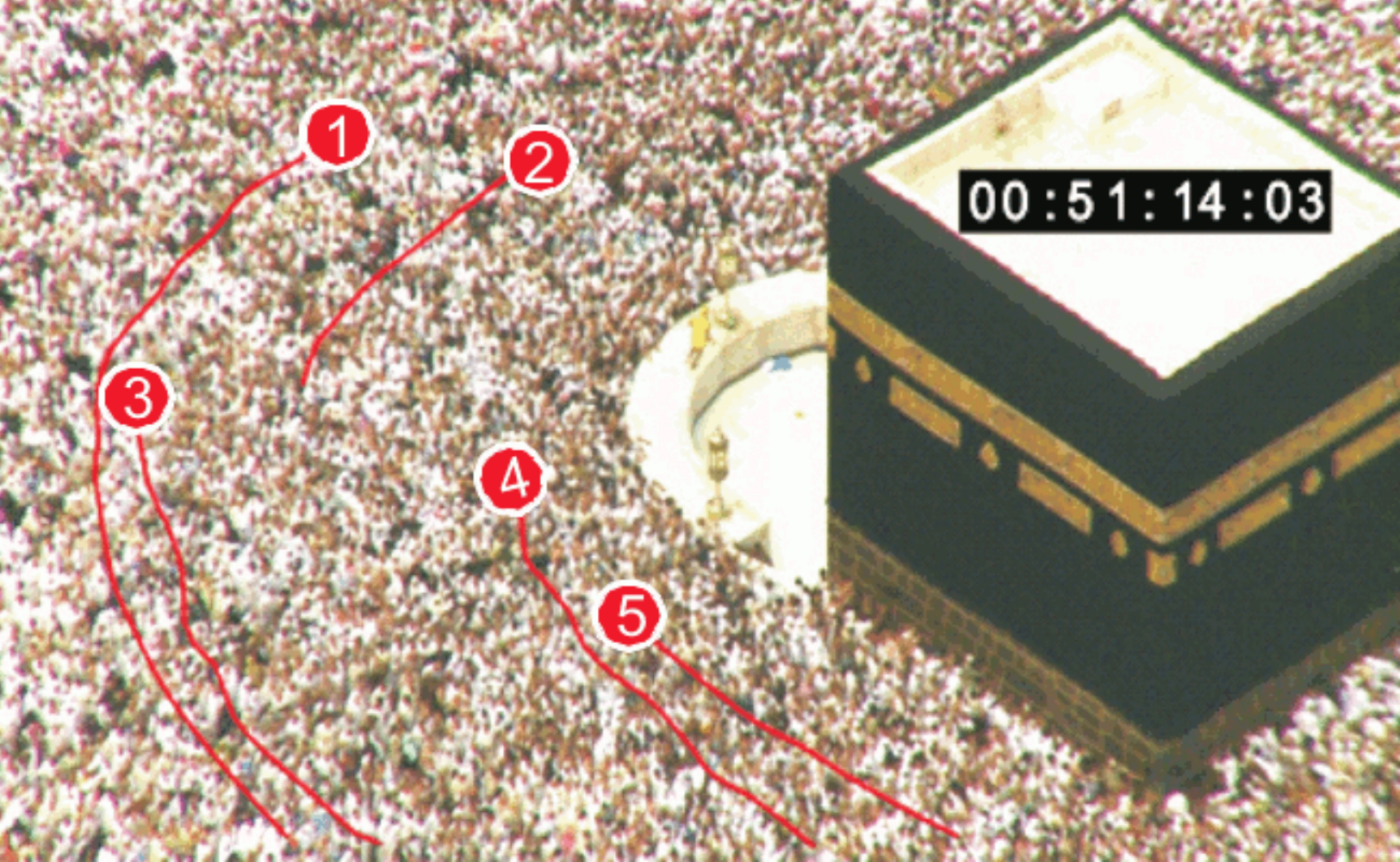}
\end{center}
\caption{Pilgrims paths. With a new computer
algorithm developed during this research, the trajectories or movements of pedestrians across the 
infrastructure over time are determined. Microscopic pedestrian fields require large 
amounts of trajectory data of individual 
pedestrians. Every red solid curve corresponds to one pedestrian trajectory. The oscillation in the pilgrims paths results from the huge pedestrian forces acting on every individual in the crowd.}
\label{fig:pilgrimspaths-2}
\end{figure}

In the literature, there is a large number of approaches on detection and tracking of moving objects from video images.
Spatio-temporal analysis has, in the past, been used to recognize walking persons, where subspaces in the video are treated as spatio-temporal volumes \cite{Ricquebourg2000}. Application of a Fourier transform to this data can then identify data relating to movement across the volume. 
This approach allowed pedestrian trajectories to be reconstructed from video with high precision, taking advantage from the methods and the high developed computational technology.
The common approach to detect movement is to produce comparison images (an image representing the different details between two images) since this is computationally efficient \cite{Masoud2001}. These comparison images can then be computed further to estimate movement vectors that describe the motion of drop-shaped objects captured in the respective images. 
Murakami and Wada demonstrate another method, filing the difference frame, and instead compare the properties of drops identified in consecutive frames \cite{Murakami}. A drop that is close to the position of a drop in a previous frame, and shares similar dimensions, is likely to refer to the same figure. Motion vectors are also used to find drop segmentation, which are subsequently merged or separated for the purpose of analysis. The same approach is applied to a 2D image to determine movement in 3D space.
Extrapolating the movement of pedestrians in 3D space from a 2D image allows for a far greater understanding of the interactions between entities, but does require exceptional calibrations of equipment for complete accuracy. The Murakami and Wada approach can be used to analyse low-quality video streams due to the frame-differencing algorithm and some trigonometry. Determining 3D motion does require precise knowledge of the angle and position of the camera, in addition to the basic topology of the scene being analysed. But 2D paths are easy to identify without these details, (see fig. \ref{fig:pilgrimspaths-2}).

In figure \ref{fig:pilgrimspaths-2} we show the path of individuals within the crowd. One clearly recognizes that the movement around the Kaaba is not a circle movement. The tracking of a single individual in the pilgrim stream indicates some oscillation movement around the main path of the individual. It is caused by the physical repulsive and attractive forces acting on the individual. Physical forces become important when an individual comes into physical contact with another individual/obstacle.  
When a local density of 6 persons per square meter is exceeded, free movement is impeded and local flow decreases, causing the outflow to drop significantly below the inflow. This causes a higher and higher compression in the crowd, until the local densities become critical in specific places on the Mataf platform.

\section{Analysis of the Pilgrims movement on the Mataf}
In the Mataf everything is dense and we have a compact state. The pilgrims have body contact in all directions and no influence on their movement; they float in the stream. This forms structures and turbulences in the flow. These turbulences can be well observed in our video recording. Density and velocity can also be seen.
These observed Hajj rituals, especially the Mataf, showed some critical points in the motion of the pilgrims that we had not paid much attention to before. For example: the edge effect, density effect, shock-wave effect etc., and phenomena like these influence the restraint of the motion and are very important to be considered.

Our video analysis shows that the pedestrian density decreases with the distance from the Kaaba wall, cf.\  figures \ref{fig:densitydistribution-5}, \ref{fig:density-1}, and \ref{fig:densitymap-1}. It is the same as the real behaviour of pilgrims on the Mataf ritual (all pilgrims want to be near to the Kaaba wall). 
Our video analysis about the Mataf area indicates that, even at extreme densities, the average local speeds and flows stay limited. This extremely high local density causes forward and backward moving shock-waves, which could be clearly observed in our video.
We can see a kind of oscillation on the pilgrims paths around the Kaaba, this oscillation is caused by shock-waves and is affected by the repulsive forces between the pedestrians in high density crowds (see fig. \ref{fig:pilgrimspaths-2}).

\section{Conclusion and possible improvement}
One of the significant challenges in the planning, design and management of public facilities subject to high density crowd dynamics and pedestrian traffic are the shortcoming in the empirical data. The collected data concerning crowd behaviour using different techniques (image processing) and analysis of ordered image sequences obtained from video recording is increasingly desirable in the design of facilities and long-term site management. We have investigated the efficiency of a number of techniques developed for crowd density estimation, movement estimation, critical places and events detection  using image processing. In the above sections and within this investigation we have presented techniques for background generation and calibration to improve the previously developed simulation model.

Even though extracting information about human characteristics from video recording may still be in its infancy, it is important to mention that the field of human motion analysis is large and has a history traced back to the work of Hoffman and Flinchbaugh \cite{Flinchbaugh1982}.
In the field of pedestrian detection techniques, moreover in the big area of computer vision, many problems have accumulated. In the human motion analysis, and also in the problem of the  detection of moving objects, remain other problems, namely to recognize, categorize, or analyse the long-term pattern of motion. The inspection of the literature in the last decade indicates increasing interest in event detection, video tracking, object recognition, because of the clear application of these technologies to problems in surveillance. 
Recently many methods have been developed to extract information about moving object like speed and density.
Almost all these systems require complex intermediate processes, such as reference points on the tracked objects or the image segmentation. One limitation of this current system is that the detection failures for these intermediates will lead to failure for the entire system.

Improvement of an algorithm to be able to reproduce traffic flow and to help in  the microscopic pedestrian data collection is very essential. Moreover the automatic video data collection will highly enhance the achievement of a system for higher pedestrian traffic densities.

\section{Acknowledgements}
I would like to express my sincerest thanks and gratitude to Prof. Dr. G. Wunner for a critical reading of the 
manuscript, for his important comments and suggestions to improve the manuscript. Many thanks to Dr. H. Cartarius for his support during writing this work.
\bibliographystyle{unsrt}
%\bibliography{literatur}

\begin{thebibliography}{10}

\bibitem{Predtetschenski-Milinski1969}
W.~M. Predtechensky and A.~I. Milinski.
\newblock {\em Personenstr{\"o}me in Geb{\"a}uden}.
\newblock Staatsverlag der Deutschen Demokratischen Republik, Berlin, russ:
  1969, germ: 1971.

\bibitem{Knoblauch1996}
R.~L. Knoblauch, M.~T. Pietrucha, and M.~Nitzburg.
\newblock Field studies of pedestrian walking speed and start-up time.
\newblock {\em Transportation Research Record 1538. Washington (DC): National
  Research Council, Transportation Research Board}, Dec:27--38, 1996.

\bibitem{Fruin1971a}
J.~J. Fruin, American~Society of~Mechanical~Engineers, and American~Society
  of~Mechanical Engineers. Standing Committee~on Transportation.
\newblock {\em Designing for Pedestrians: A Level of Service Concept}.
\newblock Univ. Microfilm, 1970.

\bibitem{o1996transport}
C.~O'Flaherty.
\newblock {\em Transport Planning and Traffic Engineering}.
\newblock Engineering village. Taylor \& Francis, 1996.

\bibitem{Fruin1987}
J.~J. Fruin.
\newblock Pedestrian planning and design.
\newblock {\em Elevator World.}, 1987.

\bibitem{Seyfried2009}
U.~Chattaraj, A.~Seyfried, and P.~Chakroborty.
\newblock {\em Comparison of pedestrian fundamental diagram across cultures}.
\newblock 2009.

\bibitem{Bulpitt1998}
A.~J. Bulpitt and N.~Sumpter.
\newblock Learning spatio-temporal patterns for predicting object behaviour.
\newblock {\em In BMVC.}, 1982.

\bibitem{heisele1998}
B.~Heisele and C.~Woehler.
\newblock Motion-based recognition of pedestrians.
\newblock {\em Proceedings Fourteenth International Conference on Pattern.},
  2:1325--30, 1998.

\bibitem{Tsuchikawa1995}
A.~Sato, H.~Koike, A.~Tomono, and M.~Tsuchikawa.
\newblock A moving-object extraction method robust against illumination level
  changes for a pedestrian counting system.
\newblock {\em Proceedings International Symposium on Computer Vision.}, 42,
  Issue 3:563--568, 1995.

\bibitem{Hoogendoorn2005}
S.~P. Hoogendoorn and W.~Daamen.
\newblock Pedestrian behavior at bottlenecks.
\newblock {\em Transportation Science}, 39 (2):147--159, 2005.

\bibitem{Johansson2008}
A.~Johansson and D.~Helbing.
\newblock From crowd dynamics to crowd safety: A video-based analysis.
\newblock {\em Advances in Complex Systems}, 4 (4):497--527, 2008.

\bibitem{Boltes2010}
M.~Boltes, A.~Seyfried, B.~Steffen, and A.~Schadschneider.
\newblock Automatic extraction of pedestrian trajectories from video
  recordings.
\newblock {\em Pedestrian and Evacuation Dynamics, Springer-Verlag Berlin
  Heidelberg}, pages 43--54, 2010.

\bibitem{Papageurgiou1999}
C.~Papageorgiou and T.~Poggio.
\newblock Trainable pedestrian detection.
\newblock {\em Proceedings 1999 International Conference on Image Processing},
  4:35--9, 1999.

\bibitem{Marana1998}
A.~N. Marana, L.~F. Costa, R.~A. Lotufo, and S.~A. Velastin.
\newblock On the efficacy of texture analysis for crowd monitoring.
\newblock {\em SIBGRAPI'98 1998, Proceedings}, pages 354--61, 1998.

\bibitem{Zhang2012}
Z.~Zhang and M.~Li.
\newblock Crowd density estimation based on statistical analysis of local
  intra-crowd motions for public area surveillance.
\newblock {\em Optical Engineering}, 51(4), 047204, 2012.

\bibitem{Verona2001}
V.~Verona and A.~Marana.
\newblock Wavelet packet analysis for crowd density estimation.
\newblock {\em in Proc. of the IASTED International Symposium on Applied
  Informatics, Cancun, Mexico}, 2001.

\bibitem{Li2006}
X.~Li, L.~Shen, and H.~Li.
\newblock Estimation of crowd density based on wavelet and support vector
  machine.
\newblock {\em Trans. Inst. Meas. Control (London)}, 28(3):299--308, 2006.

\bibitem{Wu2006}
X.~Wu.
\newblock Crowd density estimation using texture analysis and learning.
\newblock {\em in Proc. of IEEE Conf. on Robotics and Biometics, IEEE, Kunming,
  China}, 2006.

\bibitem{Sen2009}
G.~Sen, L.~Wei, and Y.~H. Ping.
\newblock Counting people in crowd open scene based on grey level dependence
  matrix.
\newblock {\em in Proc. of Intl. Conf. on Information and Automation, IEEE,
  Canada}, 2009.

\bibitem{Helbing2007}
D.~Helbing and A.~Johansson.
\newblock The dynamics of crowd disasters: An empirical study.
\newblock {\em arXiv:physics/0701203v2[physics.soc-ph]}, 2007.

\bibitem{Seyfried2005}
A.~Seyfried, B.~Steffen, W.~Klingsch, and M.~Boltes.
\newblock The fundamental diagram of pedestrian movement revisited.
\newblock {\em J. Stat. Mech.}, page 10002, 2005.

\bibitem{Lam2000}
W.~H.~K. Lam and C.~Y. Cheung.
\newblock Pedestrian speed-flow relationships for walking facilities in
  hong-kong.
\newblock {\em Journal of Transportation Engineering, ASCE}, 126(4):343--349,
  2000.

\bibitem{Weidmann1993}
U.~Weidmann.
\newblock Transporttechnik der fussg{\"a}nger.
\newblock {\em Technical Report Schriftenreihe des IVT Nr. 90, Institut für
  Verkehrsplanung, Transporttechnik, Strassen- und Eisenbahnbau, ETH
  Z{\"u}rich, Zweite, erg{\"a}nzte Auflage}, 1993.

\bibitem{Ricquebourg2000}
Y.~Ricquebourg and P.~Bouthemy.
\newblock Real-time human figure control using tracked blobs.
\newblock {\em IEEE Transactions on Pattern Analysis and Machine Intelligence},
  22(8):797--808, 2000.

\bibitem{Masoud2001}
O.~Masoud and N.~P. Papanikolopoulos.
\newblock A novel method for tracking and counting pedestrians in real-time
  using a single camera.
\newblock {\em IEEE Transactions on Vehicular Technology}, 50(5):1267--78,
  Sept. 2001.

\bibitem{Murakami}
S.~Murakami and A.~Wada.
\newblock An automatic extraction and display method of walking persons'
  trajectories.
\newblock {\em Proceedings 15th International Conference on Pattern
  Recognition}, 4:611--14, Sept. 2000.

\bibitem{Flinchbaugh1982}
D.~D. Hoffman and B.~E. Flinchbaugh.
\newblock The interpretation of biological motion.
\newblock {\em Biological Cybernetics.}, 42, Issue 3:195--204, 1982.

\end{thebibliography}

\end{multicols}
\end{document}